\documentclass{article}

\usepackage{times}
\usepackage{natbib}
\usepackage{amsmath}
\usepackage{amssymb}
\usepackage[pdftex]{graphicx}
\usepackage{booktabs}
\usepackage{hyperref}
\usepackage{float}
\hypersetup{
    colorlinks,%
    citecolor=black,%
    filecolor=black,%
    linkcolor=black,%
    urlcolor=black
}

\graphicspath{{./prototype/outputs/}{./images/}}
\bibpunct{[}{]}{;}{n}{,}{,}

\newcommand{\Ncal}{\mathcal{N}}

\begin{document}
\setlength{\parindent}{0in}
\setlength{\parskip}{0.5em}
\author{Behnam Asadi\\ \texttt{behnam.asadi@gmail.com}}
\title{Dimension-Calibrated Unexplained Mass: An Interpretable Drift
Statistic for Contamination Monitoring in Data Streams}
\date{\today}
\maketitle

\begin{abstract}
Drift detectors that work tend not to explain themselves, and drift detectors that
explain themselves tend to fail in high dimension. We close that gap for Gaussian
mixture models (GMMs): each fitted component is a named ``regime,'' and the
fraction of a stream window matching \emph{no} regime---its \emph{unexplained
mass}---is a drift signal that is simultaneously its own explanation. We identify
why this statistic collapses in high dimension and repair it. Under a correct
component a normal point in $d$ dimensions lies $\approx\!\sqrt{d}\,\sigma$ from
the mean, so once $d\gtrsim9$ essentially every point exceeds a fixed $3\sigma$
radius: window-level ROC-AUC is exactly $0.50$ on Satellite ($d{=}36$) and
Optdigits ($d{=}64$). Calibrating the radius to $\sqrt{\chi^2_d(0.99)}$ removes
the collapse---AUC $\mathbf{1.00}$ and $\mathbf{0.89}$---while leaving low
dimensions unchanged. Across seven public benchmarks, five seeds, and eight
model-free detectors spanning the kernel, classifier, projection,
density-difference, transport, likelihood and partition families, the repaired
statistic is best or tied-best on five of seven datasets at $10\%$ window
contamination (its two losses are Pendigits, where the whole field beats it,
and Optdigits), and as
contamination becomes sparse the sample-level detectors fade toward chance
while it degrades most gracefully: at $2\%$ its mean AUC across the benchmarks
is $0.86$ against at most $0.73$ for any model-free detector ($1.00$ vs.\
MMD's $0.72$ on KDD-http)---while alone among them reporting \emph{which}
regime the data left and \emph{how far} outside it the window lies. We delimit its scope honestly:
unexplained mass detects and explains \emph{novel-regime} drift but is blind by
construction to \emph{in-support} re-weighting of known regimes, where
distribution-level tests are required and explain nothing; and the underlying
density model's EVT-calibrated false-alarm rates degrade above $d\approx36$.
All code and experiments are released.
\end{abstract}


\section{Introduction}\label{Introduction}
\begin{figure}[H]
\centering
\includegraphics[scale=0.3]{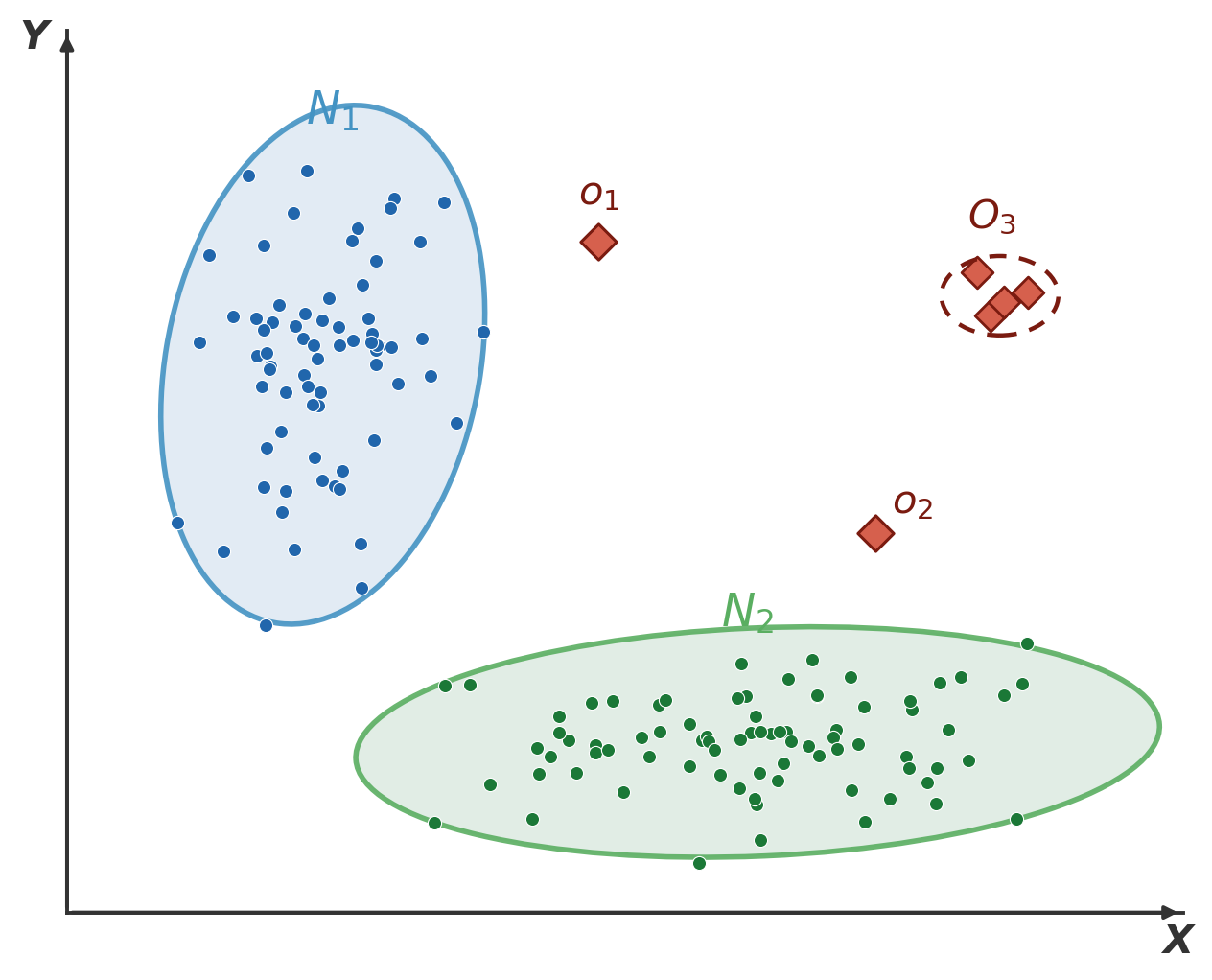}
\caption{A simple example of anomalies in a two-dimensional data set: $N_1,N_2$
are normal regions; $o_1,o_2$ are point anomalies; $O_3$ is a small anomalous
cluster.}
\label{fig:illustration}
\end{figure}

Anomalies are patterns in data that do not conform to a well-defined notion of
normal behaviour. Detecting them is critical across domains: intrusion detection
for cyber-security, fraud detection in financial transactions, fault detection
in safety-critical systems, and condition monitoring of machines and robots
\citep{Chandola2009}. Figure~\ref{fig:illustration} illustrates the idea in two
dimensions: most observations fall in the dense regions $N_1$ and $N_2$, whereas
points $o_1$, $o_2$ and the small cluster $O_3$ lie far from them and are
flagged as anomalous.

This work originated in monitoring the joint-current streams of a six-legged
walking robot, but the method is domain-agnostic and we evaluate it as such. We
draw a distinction that is often blurred in practice and that organises the
whole approach:
\begin{itemize}
\item \textbf{Point anomalies:} is an \emph{individual} observation unlikely
under normal behaviour? This is naturally answered by the likelihood a density
model assigns to the point.
\item \textbf{Distributional / drift anomalies:} has the \emph{distribution} of
a recent window of the stream shifted away from normal behaviour? Individual
points may each look normal while their collective distribution has changed---a
regime change, sensor degradation, or a novel operating mode
\citep{Gama2014,Pimentel2014}. This is naturally answered by a
\emph{distance between distributions}.
\end{itemize}
Gaussian Mixture Models serve both tasks with a single, interpretable model: a
weighted set of Gaussian components that approximates any continuous density to
arbitrary accuracy \citep{Bishop2006,Reynolds2009}. Unlike black-box detectors,
a GMM exposes named ``regimes'' (its components), so an alarm can be attributed
to \emph{which} regime the data left and \emph{how far}. The catch---and the open
problem this paper addresses---is that the natural interpretable drift statistic on
such a model, the fraction of a window lying outside every regime, silently loses
all resolution as dimension grows; we show the failure is a simple concentration
effect and that calibrating the regime radius to the dimension repairs it
completely.

\paragraph{Contributions.}
(i) \textbf{A diagnosis.} We identify the concentration mechanism by which the
fixed-radius unexplained-mass statistic fails in high dimension---under a correct
component a normal point lies at $\approx\!\sqrt{d}\,\sigma$, so a fixed $3\sigma$
radius is exceeded by essentially every point once $d\gtrsim9$---and exhibit it as an exact collapse to
chance AUC at $d{=}36$ and $d{=}64$.
(ii) \textbf{A fix.} Calibrating the radius to the data dimension via the
$\chi^2_d(0.99)$ quantile---a classical prescription from multivariate outlier
detection that, to our knowledge, has not been applied to window-level drift
statistics on mixture models---removes the saturation entirely (window-AUC
$0.50\to1.00$ on Satellite, $0.50\to0.89$ on Optdigits, low dimensions unchanged),
yielding an \emph{interpretable} drift statistic that performs on par with
eight model-free detectors spanning the kernel, classifier, projection,
density-difference, transport, likelihood and partition families of the recent
drift literature across $d\in[3,64]$, and degrades most gracefully of all
detectors tested as contamination becomes sparse (mean AUC $0.86$ vs.\ at most
$0.73$ at a $2\%$ rate).
(iii) \textbf{A scope characterisation.} We separate \emph{contamination} drift
from \emph{in-support} regime re-weighting and show the dissociation is structural,
not incidental: the unexplained-mass statistic detects and explains the first and
is blind to the second, while MMD detects both and explains neither. This tells a
practitioner which detector a given drift regime requires.
(iv) \textbf{A reproducible pipeline and an honest benchmark.} We couple automatic
model selection (BIC, $k$-means-initialised) with likelihood-based point scoring and
Extreme-Value-Theory thresholds, and report measured versus target false-alarm
rates, seven public benchmarks ($d\in[3,64]$), six point-anomaly baselines,
eight model-free drift baselines, and five random splits---including where the
GMM loses. All code is released.

\section{Related Work}\label{RelatedWork}
Anomaly and novelty detection are surveyed in \citet{Chandola2009} and
\citet{Pimentel2014}. GMMs have long been used for density-based anomaly
detection \citep{Reynolds2009}; the parameters are estimated by
Expectation--Maximization \citep{Dempster1977}, and the number of components is
commonly chosen by the Bayesian Information Criterion \citep{Schwarz1978} or the
Integrated Completed Likelihood \citep{Biernacki2000}. Popular alternative
detectors include Isolation Forest \citep{Liu2008}, the Local Outlier Factor
\citep{Breunig2000}, and the one-class SVM \citep{Scholkopf2001}; more recent
tabular detectors include the parameter-free ECOD \citep{Li2022ECOD} and COPOD
\citep{Li2020COPOD}, and autoencoder reconstruction-error methods
\citep{Sakurada2014}; the ADBench benchmark \citep{han2022adbench} compares
thirty such algorithms at scale and finds, consistent with our results, that no
detector dominates. We benchmark against all of the above. None of these,
however, exposes an interpretable model of normality.

Comparing two GMMs requires a distance between mixtures. The KL divergence,
central to statistical pattern recognition \citep{Fukunaga1990}, has \emph{no}
closed form between two Gaussian mixtures; practical work therefore uses
approximations---Monte-Carlo, the variational and matching bounds of
\citet{Hershey2007}, or the matching approximation of \citet{Goldberger2003}.
An alternative is to use a divergence that \emph{is} closed-form for mixtures:
\citet{Kampa2011} derived such an expression for the Cauchy--Schwarz divergence,
which we adopt. Distance-based comparison of GMMs has been used, e.g., for
speaker and music similarity \citep{Jensen2007}. For streaming settings,
Extreme Value Theory provides principled thresholds at a chosen false-alarm rate
\citep{Coles2001,Siffer2017}.

Drift detection itself is a mature field, surveyed classically by
\citet{Gama2014} and, for the modern period, by \citet{lu2019learning},
\citet{gemaque2020overview} for the unsupervised setting, and
\citet{hinder2024onePartA}; \citet{lukats2025benchmark} benchmark fully
unsupervised detectors under a shared protocol. Since our setting provides no
labels at monitoring time, the relevant family is \emph{unsupervised,
multivariate} detection, where the main strategies are: comparing windows
through a learned discriminator (the classifier two-sample test of
\citet{lopezpaz2017revisiting}, deployed for streams as D3 by
\citet{gozuacik2019d3}); comparing windows through a distribution distance,
either directly (kernel MMD \citep{Gretton2012}, its learned-kernel
\citep{liu2020deepkernel}, aggregated \citep{schrab2023mmdagg}, and automated
\citep{kubler2022automl} refinements, density-difference estimation
\citep{sugiyama2013lsdd,bu2018lsdd}, transport distances \citep{rabin2011sliced})
or after a projection or partition (principal components
\citep{goldenberg2019survey,goldenberg2020pca}, equal-intensity partitions
\citep{liu2021eikmeans}, image encodings \citep{desouza2021ibdd});
model-based statistics such as the semi-parametric log-likelihood of
\citet{kuncheva2013spll}---the closest ancestor of our statistic, scoring a
window by its Mahalanobis fit to reference clusters---or independence-with-time
tests \citep{hinder2020dawidd}; and monitors coupled to a deployed model
\citep{cerqueira2023studd,assis2025adwinu}, to learned representations
\citep{wan2024mcddd,greco2025driftlens}, or to meta-feature summaries of
concepts \citep{halstead2023combining}. Dimensionality-reduction plus
two-sample-test pipelines were studied empirically by \citet{rabanser2019failing},
and \citet{liu2023cdddi} argue that threshold selection, not the choice of
statistic, often dominates detector behaviour---precisely the failure mode our
$\chi^2_d$ calibration addresses. Section~\ref{Experiments} compares against
representatives of each model-free family (C2ST, SPLL, PCA-based, LSDD, sliced
Wasserstein) alongside MMD.

Interpretable and explainable drift detection is an active area in its own right:
\citet{Pelosi2025} survey it systematically, \citet{hinder2024onePartB} review
drift localisation and explanation, and specific mechanisms include model-based
explanations \citep{hinder2023modelbased} and local attribution in streams
\citep{haug2022cdleeds}. The thresholding fact we exploit is
classical: under a correct Gaussian component the squared Mahalanobis distance is
$\chi^2_d$-distributed, a standard prescription for multivariate outlier flagging
\citep{Fukunaga1990}. Our contribution is not that fact but its consequence for
drift monitoring on mixtures: the customary fixed $3\sigma$ regime radius silently
destroys the window-level unexplained-mass statistic in high dimension, and the
$\chi^2_d$ quantile restores it to parity with a kernel two-sample test while
keeping the per-regime explanation.

\section{Method}\label{Method}

\subsection{Gaussian mixture density model}
A GMM models a $d$-dimensional observation $x$ as a weighted sum of $K$ Gaussian
components,
\begin{equation}\label{eq:gmm}
p(x\mid\lambda)=\sum_{k=1}^{K}\pi_k\,\Ncal(x\mid\mu_k,\Sigma_k),
\qquad \sum_{k=1}^{K}\pi_k=1,\ \pi_k\ge 0,
\end{equation}
with parameters $\lambda=\{\pi_k,\mu_k,\Sigma_k\}_{k=1}^K$ estimated by
Expectation--Maximization \citep{Dempster1977,Bishop2006}. Each component is a
regime of normal behaviour; the weight $\pi_k$ is the prior probability of that
regime.

\subsection{Model selection and initialisation}
The number of components $K$ is not known a priori: too few yields an
under-fitting density, too many increases variance and cost. Rather than fixing
$K$, we select it by the Bayesian Information Criterion
\begin{equation}\label{eq:bic}
\mathrm{BIC}(K)=-2\log \hat{L}(K)+p_K\log N,
\end{equation}
where $\hat L(K)$ is the maximised likelihood, $p_K$ the number of free
parameters and $N$ the sample size, choosing $K^\star=\arg\min_K\mathrm{BIC}(K)$.
BIC is the natural criterion here because, unlike hard-clustering validity
indices, it scores the mixture \emph{likelihood} directly. EM is sensitive to
initialisation, so we seed it with $k$-means: cluster centroids initialise the
means $\mu_k$, and the fraction of points per cluster initialises the weights
$\pi_k$.

\subsection{Point-anomaly scoring}
Given a GMM fitted to normal data, an observation $x$ is scored by its negative
log-likelihood
\begin{equation}\label{eq:nll}
s(x)=-\log\sum_{k=1}^{K}\pi_k\,\Ncal(x\mid\mu_k,\Sigma_k),
\end{equation}
so that unlikely points receive high scores. This is a per-sample test suited to
point anomalies.

\subsection{Detecting distributional drift}\label{sec:drift}
For a window $W=\{x_1,\dots,x_m\}$ of the stream we ask whether its distribution has
departed from the reference normal GMM $f$. Our primary, \emph{interpretable}
statistic is the \emph{unexplained mass}: the fraction of the window that matches no
reference regime beyond $\tau$ standard deviations,
\begin{equation}\label{eq:unexplained}
U_\tau(W)=\frac{1}{m}\sum_{i=1}^{m}\mathbf{1}\!\left[\min_{k}\Delta_k(x_i)>\tau\right],
\qquad \Delta_k(x)=\sqrt{(x-\mu_k)^\top\Sigma_k^{-1}(x-\mu_k)},
\end{equation}
where $\Delta_k$ is the Mahalanobis distance to component $k$. $U_\tau$ rises when a
window contains points outside every known regime, and the value \emph{is} the
explanation. We evaluate two thresholds: the customary fixed radius $\tau=3$
(the univariate three-sigma convention of statistical process control
\citep{shewhart1931economic}, carried unchanged into multivariate practice),
and the \emph{dimension-calibrated} radius
\begin{equation}\label{eq:taud}
\tau_d=\sqrt{\chi^2_d(0.99)},
\end{equation}
i.e.\ a point is unexplained when its \emph{squared} distance
$\min_k\Delta_k^2(x)$ exceeds the $0.99$ quantile of the $\chi^2_d$
distribution ($\tau_d\approx4.1$ at $d{=}6$, $\approx9.7$ at $d{=}64$; note the
quantile calibrates the squared distance, so the radius is its square root).
As a control for the window statistic itself we also score each window by its
mean negative log-likelihood under the reference GMM---the generic aggregation
of the same model's point-anomaly score, with no per-point threshold and no
regime attribution.
Section~\ref{Experiments} shows the fixed radius silently fails in high
dimension and the calibrated one repairs it. As a closed-form, model-level alternative we also
fit a GMM $g$ to the window and measure a GMM-to-GMM divergence to $f$. The KL
divergence
\begin{equation}\label{eq:kl}
D_{\mathrm{KL}}(f\Vert g)=\int f(x)\log\frac{f(x)}{g(x)}\,\mathrm{d}x
\end{equation}
is closed-form for two single Gaussians but \emph{not} for two mixtures; it is
also asymmetric and unbounded, which makes it awkward as a threshold statistic.
We therefore use the \emph{Cauchy--Schwarz divergence} \citep{Kampa2011},
\begin{equation}\label{eq:cs}
D_{\mathrm{CS}}(f,g)=-\log\!\left(
\frac{\int f(x)g(x)\,\mathrm{d}x}
     {\sqrt{\int f(x)^2\,\mathrm{d}x\ \int g(x)^2\,\mathrm{d}x}}\right),
\end{equation}
which is symmetric and satisfies $D_{\mathrm{CS}}\ge 0$ with equality iff
$f=g$. For GMMs $f=\sum_i\pi_i\Ncal(\cdot\mid\mu_i,\Sigma_i)$ and
$g=\sum_j\pi'_j\Ncal(\cdot\mid\mu'_j,\Sigma'_j)$, every integral above is a sum
of Gaussian--Gaussian inner products with the exact identity
\begin{equation}\label{eq:inner}
\int \Ncal(x\mid\mu_i,\Sigma_i)\,\Ncal(x\mid\mu'_j,\Sigma'_j)\,\mathrm{d}x
=\Ncal(\mu_i\mid\mu'_j,\ \Sigma_i+\Sigma'_j),
\end{equation}
giving the closed form
\begin{equation}\label{eq:csgmm}
D_{\mathrm{CS}}(f,g)=
-\log\!\Big(\textstyle\sum_{i,j}\pi_i\pi'_j\,z_{ij}\Big)
+\tfrac12\log\!\Big(\textstyle\sum_{i,j}\pi_i\pi_j\,z^{ff}_{ij}\Big)
+\tfrac12\log\!\Big(\textstyle\sum_{i,j}\pi'_i\pi'_j\,z^{gg}_{ij}\Big),
\end{equation}
where $z_{ij}=\Ncal(\mu_i\mid\mu'_j,\Sigma_i+\Sigma'_j)$ is the cross-term and
$z^{ff}_{ij}=\Ncal(\mu_i\mid\mu_j,\Sigma_i+\Sigma_j)$,
$z^{gg}_{ij}=\Ncal(\mu'_i\mid\mu'_j,\Sigma'_i+\Sigma'_j)$ are the self-terms of
$f$ and $g$. All sums are evaluated in the log domain for numerical stability.%
\footnote{The same Gaussian--product identity~\eqref{eq:inner} yields other
closed-form mixture divergences. Writing the mixture inner product as
$\langle f,g\rangle=\sum_{i,j}\pi_i\pi'_j\,z_{ij}$, one may replace the
Cauchy--Schwarz geometric-mean normalisation $\sqrt{\langle f,f\rangle\langle
g,g\rangle}$ with an arithmetic-mean one, giving
$-\log\!\big(2\langle f,g\rangle/(\langle f,f\rangle+\langle g,g\rangle)\big)
=-\log\!\big(1-\lVert f-g\rVert^2/(\langle f,f\rangle+\langle g,g\rangle)\big)$,
a log-normalised integrated-squared-error ($L_2$) divergence. It is likewise
symmetric, non-negative and zero iff $f=g$, coincides with $D_{\mathrm{CS}}$
when $\langle f,f\rangle=\langle g,g\rangle$, and behaves similarly in practice;
we report $D_{\mathrm{CS}}$ as the representative closed-form divergence.}
For comparison we also implement a
matching-based KL \emph{surrogate} in the spirit of \citet{Goldberger2003}
(and its symmetrised, Jeffreys form), used only as an ablation baseline.

\subsection{Thresholding with Extreme Value Theory}\label{sec:evt}
A detector needs a threshold at a controllable false-alarm rate (FAR). Fitting
on normal scores only, we use the Peaks-Over-Threshold method
\citep{Coles2001,Siffer2017}: exceedances of the normal scores over a high
quantile $u$ are modelled by a Generalized Pareto distribution, and the
threshold for a target FAR is obtained by extrapolating its tail. In principle
this extends to FARs below the empirical resolution $1/N$, where a raw quantile
has no support; we do not test that regime here and evaluate calibration at
$10^{-2}$ and $10^{-3}$ (Section~\ref{ThresholdCalib}), reporting the empirical
quantile as a baseline.

\section{Experiments}\label{Experiments}

\subsection{Datasets}
We use seven public benchmarks (Table~\ref{tab:datasets}) spanning intrusion
detection, medical imaging, aerospace telemetry and handwriting, with
dimensionality from $3$ to $64$. KDD-http/smtp come from the KDD~Cup~99 corpus
\citep{Tavallaee2009}; the remainder follow the conventions of the ODDS library
\citep{ODDS2016}. Features are $z$-scored (byte counts are $\log$-transformed
first). Anomaly-labelling conventions are documented in the released code.

\begin{table}[H]
\centering
\caption{Benchmark datasets, ordered by dimensionality.}
\label{tab:datasets}
\begin{tabular}{lrrrr}
\hline
Dataset & $n$ & $d$ & \#anom. & \%anom. \\
\hline
KDD-http & 58725 & 3 & 2209 & 3.76 \\
KDD-smtp & 95156 & 3 & 30 & 0.03 \\
Mammography & 11183 & 6 & 260 & 2.32 \\
Shuttle & 49097 & 9 & 3511 & 7.15 \\
Pendigits & 10049 & 16 & 201 & 2.00 \\
Satellite & 5100 & 36 & 75 & 1.47 \\
Optdigits & 5169 & 64 & 103 & 1.99 \\
\hline
\end{tabular}

\end{table}

\subsection{Protocol and baselines}
We follow a semi-supervised (one-class) protocol: each detector is trained on
$60\%$ of the \emph{normal} data and tested on the remaining normal data plus all
anomalies. To remove sensitivity to a particular split we repeat this over five
random splits and report ROC-AUC as mean\,$\pm$\,standard deviation (and average
precision, AP, in the released results). Baselines span classic and modern
detectors: Isolation Forest \citep{Liu2008}, Local Outlier Factor
\citep{Breunig2000}, a one-class SVM \citep{Scholkopf2001} (kernel-approximated for
scalability), the parameter-free ECOD \citep{Li2022ECOD} and COPOD
\citep{Li2020COPOD} (via PyOD \citep{Zhao2019PyOD}), and a shallow bottleneck
autoencoder scored by reconstruction error \citep{Sakurada2014}. All methods use
scikit-learn \citep{Pedregosa2011}; the GMM selects $K\in\{1,\dots,10\}$ by BIC.

\subsection{Point-anomaly detection}
\begin{figure}[H]
\centering
\includegraphics[width=\textwidth]{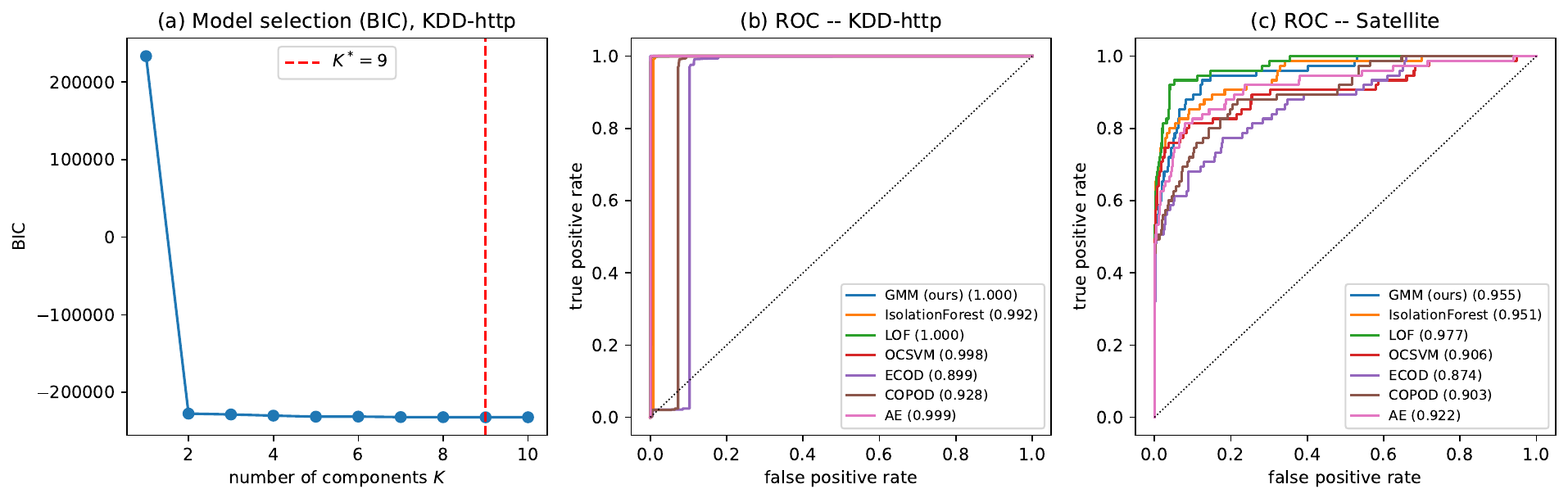}
\caption{(a) BIC model selection on KDD-http. (b,c) ROC curves for the GMM
detector and baselines on a low- and a high-dimensional benchmark.}
\label{fig:benchmark}
\end{figure}

Table~\ref{tab:benchmark} reports ROC-AUC across seven detectors. We are explicit
that the GMM is \emph{not} the most accurate: LOF is the strongest on five of seven
datasets, sometimes by a wide margin (e.g.\ Optdigits $0.97$ vs the GMM's $0.86$),
and the modern parameter-free ECOD takes the remaining column (Mammography,
$0.91$, with COPOD essentially tied). The GMM tops only KDD-http (tied). Adding these stronger and
more recent baselines does not change the picture: no single method dominates, and
the GMM's value is not the score. What it offers is \emph{competitiveness at low
cost}: it stays within a few AUC points of the best on the lower-dimensional
benchmarks, never inverts, and---unlike LOF, Isolation Forest, one-class SVM, ECOD,
COPOD or the autoencoder---yields an interpretable density model that also supports
the drift detection and regime attribution of the following sections. LOF's edge, in particular, comes at the cost of storing the
entire training set and providing no model to inspect. Table~\ref{tab:capab}
summarises this trade-off: the accuracy the GMM concedes buys capabilities the
other detectors do not provide directly (``regime-level attribution'' means
the detector's own output names the regime; post-hoc explanation methods can
of course be bolted onto any detector). We also note that our one-class SVM is a scalable
Nystr\"om\,+\,SGD approximation rather than an exact kernel SVM; its low score on
Pendigits ($0.35$) reflects that approximation and untuned bandwidth, not a
limitation of the exact one-class SVM. Accuracy also degrades with dimension:
the GMM trails LOF clearly on the $64$-dimensional Optdigits set, consistent with
the curse of dimensionality for density estimation. Figure~\ref{fig:benchmark}(a)
shows the BIC curve on KDD-http (a sharp elbow at $K{=}2$ with BIC decreasing only
marginally thereafter, so the exact $K$ is not critical); panels (b,c) show ROC
curves on an easy ($3$D) and a hard ($36$D) benchmark.

\begin{table}[H]
\centering
\caption{Point-anomaly detection, ROC-AUC (mean over five random splits; best per
dataset column in bold). Seven detectors $\times$ seven datasets; the GMM is
competitive but rarely best (LOF wins most columns), and no baseline collapses as
badly as the approximate OCSVM on Pendigits.}
\label{tab:benchmark}
\begin{tabular}{lccccccc}
\hline
Method & http & smtp & Mammo & Shut & Pen & Sat & Opt \\
\hline
GMM (ours) & \textbf{1.000} & 0.882 & 0.871 & 0.999 & 0.969 & 0.956 & 0.859 \\
IForest & 0.992 & 0.883 & 0.877 & 0.996 & 0.788 & 0.952 & 0.849 \\
LOF & 1.000 & \textbf{0.903} & 0.845 & \textbf{1.000} & \textbf{0.986} & \textbf{0.978} & \textbf{0.974} \\
OCSVM & 0.996 & 0.819 & 0.772 & 0.995 & 0.348 & 0.907 & 0.532 \\
ECOD & 0.902 & 0.880 & \textbf{0.906} & 0.993 & 0.478 & 0.876 & 0.618 \\
COPOD & 0.930 & 0.792 & 0.906 & 0.995 & 0.553 & 0.906 & 0.697 \\
AE (MLP) & 0.999 & 0.767 & 0.880 & 0.995 & 0.897 & 0.920 & 0.716 \\
\hline
\multicolumn{8}{l}{\footnotesize ROC-AUC, mean over 5 random splits (std $\leq0.09$; per-split values in \texttt{results.json}); best per}\\
\multicolumn{8}{l}{\footnotesize column in bold. Columns: http, smtp, Mammo(graphy), Shut(tle), Pen(digits), Sat(ellite), Opt(digits).}\\
\end{tabular}

\end{table}

\begin{table}[H]
\centering
\caption{What each detector offers beyond an accuracy number. The GMM trades a
little accuracy (Table~\ref{tab:benchmark}) for capabilities the others lack.}
\label{tab:capab}
\begin{tabular}{lcccc}
\hline
Method & Interpretable & Density & Regime-level & Compact model \\
 & regimes & model & attribution & (no data kept) \\
\hline
GMM (ours) & \checkmark & \checkmark & \checkmark & \checkmark \\
IsolationForest & -- & -- & -- & \checkmark \\
LOF & -- & -- & -- & stores all points \\
OCSVM & -- & -- & -- & partial \\
ECOD / COPOD & per-feature & -- & -- & \checkmark \\
AE (MLP) & -- & -- & -- & \checkmark \\
\hline
\end{tabular}

\end{table}

\subsection{Threshold calibration}\label{ThresholdCalib}
A detector is only usable with a threshold at a controllable false-alarm rate
(FAR). Table~\ref{tab:evt} evaluates the EVT/Peaks-Over-Threshold thresholds of
Section~\ref{sec:evt}: we set a target FAR on normal training scores and measure
the FAR actually achieved on held-out data. On the lower-dimensional benchmarks
the achieved FAR closely tracks the target (e.g.\ KDD-http, KDD-smtp,
Mammography, Shuttle within a factor of ${\sim}1$--$2$ of the target at $10^{-2}$;
at $10^{-3}$ three of the four stay at or below target---KDD-http
$\sim\!10\times$ below---while Mammography overshoots by ${\sim}1.7\times$), and
detection rates are high where the density model is accurate. On
the highest-dimensional sets (Satellite, Optdigits) the achieved FAR drifts above
target, an honest symptom of GMM density estimation degrading in high dimensions
(Section~\ref{Discussion}). EVT thus delivers calibrated thresholds precisely in
the low-to-moderate-dimensional regime where the GMM is appropriate.

\begin{table}[H]
\centering
\caption{EVT threshold calibration: target vs.\ achieved false-alarm rate (FAR)
for the GMM detector.}
\label{tab:evt}
\begin{tabular}{lcccc}
\hline
Dataset & \multicolumn{2}{c}{target FAR $=10^{-2}$} & \multicolumn{2}{c}{target FAR $=10^{-3}$} \\
 & achieved FPR & detect.\ & achieved FPR & detect.\ \\
\hline
KDD-http & 0.0096 & 1.00 & 0.0001 & 1.00 \\
KDD-smtp & 0.0106 & 0.68 & 0.0007 & 0.67 \\
Mammography & 0.0108 & 0.34 & 0.0016 & 0.20 \\
Shuttle & 0.0114 & 0.99 & 0.0009 & 0.96 \\
Pendigits & 0.0112 & 0.19 & 0.0016 & 0.01 \\
Satellite & 0.0178 & 0.61 & 0.0038 & 0.51 \\
Optdigits & 0.0413 & 0.15 & 0.0121 & 0.01 \\
\hline
\multicolumn{5}{l}{\footnotesize Mean over five splits; ``detect.'' is the fraction of anomalies flagged (TPR).}\\
\end{tabular}

\end{table}

\subsection{Distributional drift detection}
\begin{figure}[H]
\centering
\includegraphics[width=\textwidth]{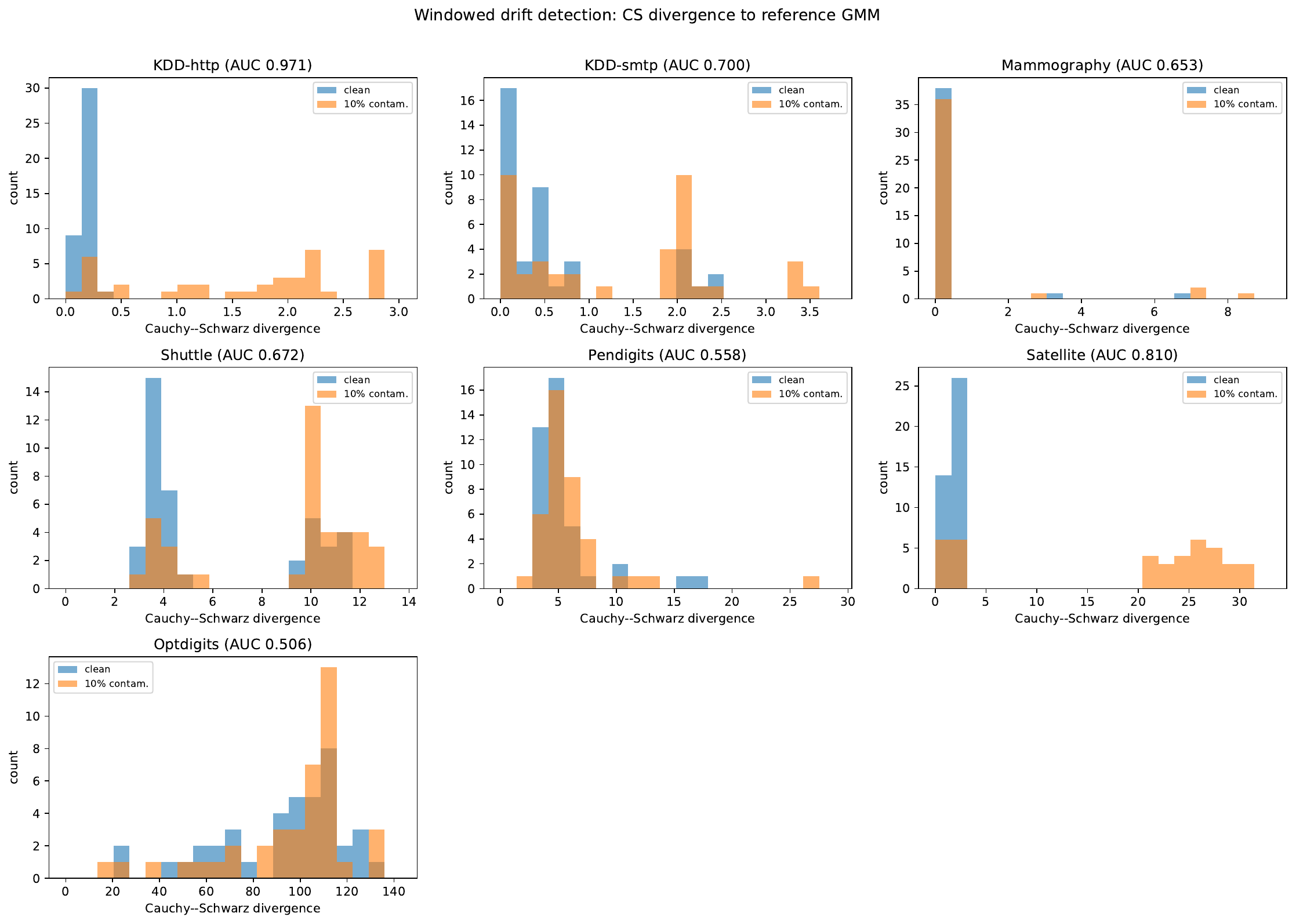}
\caption{Cauchy--Schwarz divergence from each window's GMM to the reference
normal GMM: clean vs.\ $10\%$-contaminated windows.}
\label{fig:drift}
\end{figure}

We fit a reference GMM on normal data and, for each dataset and each of five
seeds, draw $40$ clean windows and $40$ windows contaminated with $10\%$
anomalies, each window holding $500$ points (capped at a quarter of the clean
test pool); each window is modelled by its own GMM and compared to the reference by
the Cauchy--Schwarz divergence. Contaminated windows have larger divergence than
clean ones, with window-level separation above chance on all but the
$64$-dimensional Optdigits set (CS window-AUC $0.51$--$0.97$; highest on KDD-http
at $0.97$, effectively chance on Optdigits at $0.51$; Table~\ref{tab:driftcmp}).
Figure~\ref{fig:drift} shows the corresponding divergence histograms. We note the
``clean'' divergence is sizeable on some datasets (e.g.\ Shuttle, Pendigits)
because a $500$-point window GMM is itself a noisy refit, so part of the statistic
is fit noise rather than true drift. The comparison below places CS against
stronger alternatives.

\paragraph{Drift detectors compared.}
Table~\ref{tab:driftcmp} compares four window-level statistics: the exact
Cauchy--Schwarz (CS) divergence; a matching-based KL surrogate; the
\emph{unexplained-mass} statistic $U_\tau$ (Eq.~\ref{eq:unexplained}, the
interpretable GMM detector, i.e.\ the fraction of a window beyond $3\sigma$ of
every reference regime); and, as a model-free reference, the kernel Maximum Mean
Discrepancy (MMD) two-sample test \citep{Gretton2012}, which compares the window
sample directly to a reference normal sample without fitting any mixture.
Table~\ref{tab:driftbase} then places the calibrated statistic against seven
further model-free detectors spanning the main families of the recent
unsupervised-drift literature (Section~\ref{RelatedWork}): the
\emph{classifier two-sample test} (C2ST) \citep{lopezpaz2017revisiting},
deployed for streams as the D3 detector \citep{gozuacik2019d3}, scored as the
out-of-fold ROC-AUC of a logistic-regression
domain classifier; the \emph{semi-parametric log-likelihood} (SPLL) of
\citet{kuncheva2013spll}, the classical GMM-based change statistic (mean
min-Mahalanobis$^2$ of one sample under the other's clusters with pooled
covariance, taken as the maximum over both directions, as in the original); a
\emph{PCA-based} detector in the framework of
\citet{goldenberg2020pca}, taking the largest per-component
Kolmogorov--Smirnov statistic along reference principal components;
\emph{least-squares density difference} (LSDD)
\citep{sugiyama2013lsdd,bu2018lsdd}; the \emph{sliced Wasserstein-1
distance} \citep{rabin2011sliced}; and two partition-based detectors,
\emph{QuantTree} \citep{boracchi2018quanttree} (Pearson statistic over $K{=}32$
equal-probability random-split bins, with the PCA rotation and tie-breaking
noise of the original; its online extension QT-EWMA appears in
\citet{frittoli2023quanttree}) and \emph{EI-kMeans} \citep{liu2021eikmeans}
($\chi^2$ homogeneity test over equal-intensity $k$-means partitions, scored as
$-\log_{10}p$; its validity rules adapt the partition count per dataset).
Table~\ref{tab:impl} records the exact configuration of every detector;
all randomization is seeded, and the implementations are released with the
paper.

\begin{table}[H]
\centering
\caption{Baseline implementation details. ``Reference'' is what the detector
is fitted on: the $500$-point reference sample shared with MMD (drawn once per
seed), or the full training set for the partition detectors, whose papers
assume references of thousands of points (the reference GMM itself is fitted
on the full training set). All sample-based detectors operate on
reference-standardized features (near-constant features keep unit scale); all
tuning is fixed from the reference before any window is seen.}
\label{tab:impl}
\footnotesize
\begin{tabular}{p{0.13\textwidth}p{0.52\textwidth}p{0.25\textwidth}}
\hline
Detector & Configuration & Window score \\
\hline
MMD & RBF kernel, bandwidth by median heuristic on training data & unbiased
$\mathrm{MMD}^2$ vs.\ reference sample \\
C2ST & logistic regression (scikit-learn, \texttt{max\_iter}=1000,
class-balanced), stratified 5-fold CV & out-of-fold ROC-AUC
(reference$=$0 vs.\ window$=$1) \\
SPLL & $k$-means with $K$ = the reference GMM's BIC-selected $K$; pooled
within-cluster covariance with scale-aware ridge
$10^{-3}\,\mathrm{tr}(\Sigma)/d$; both directions, as in the original & max
of the two mean min-Mahalanobis$^2$ values \\
PCA-CD & PCA fitted on the reference sample; components covering $99.9\%$
of variance & max per-component two-sample KS statistic \\
LSDD & $300$ Gaussian centres subsampled from the reference; bandwidth by
median heuristic; ridge $\lambda=10^{-3}$ & $2h^{\top}\alpha -
\alpha^{\top}H\alpha$ \\
SW & $128$ seeded random projections & mean of $1$-D Wasserstein-$1$
distances \\
QTree & $K{=}32$ equal-probability bins (random axis, Bernoulli tail per
split); PCA rotation fitted on full training set; tie-breaking noise
$\Ncal(0,10^{-3})$ on training and test data, as in the original & Pearson
statistic vs.\ uniform bin probabilities \\
EI-kM & $K_0{=}25$ capped by the paper's validity rules (final $K$ ranged
$2$--$15$ across datasets); greedy equal-intensity medoid initialisation;
amplify--shrink search $\theta\in\{0,0.05,\dots,0.95\}$; built on full
training set & $-\log_{10}p$ of the $2\times K$ $\chi^2$ homogeneity test \\
\hline
\end{tabular}
\end{table}
The sample-based detectors operate on the same reference sample and
windows as MMD, with any tuning (kernel bandwidths, LSDD centres and ridge,
projection directions) fixed from the reference sample alone; the two
partition detectors are built on the full training set, as their papers
assume---the same information the reference GMM itself is fitted on. We do not
run four cited recent detectors because they answer a different protocol, not
because they are inconvenient: ADWIN-U \citep{assis2025adwinu} is a sequential
change-point monitor over feature statistics, not a batch two-sample score;
MCD-DD \citep{wan2024mcddd} and DriftLens \citep{greco2025driftlens} require
training contrastive or deep representations, which our tabular,
label-free-at-monitoring-time setting does not provide; and STUDD
\citep{cerqueira2023studd} monitors a deployed predictive model that does not
exist in an unsupervised benchmark. Learned-kernel MMD \citep{liu2020deepkernel}
and aggregated MMD \citep{schrab2023mmdagg} are omitted as strictly stronger
variants of the MMD column: the former requires per-window kernel training that
is unstable at our window size ($n{=}500$ split for training and testing), and
the latter returns a multiple-testing decision whose permutation-quantised
p-values are ill-suited to window-level ROC analysis.

The central finding is the failure and repair of the interpretable statistic. The
\emph{fixed} $\tau=3\sigma$ version works in low-to-moderate dimensions (AUC $1.0$
on KDD-http, KDD-smtp, Mammography, Shuttle; $0.87$ on Pendigits) but saturates in
high dimensions ($0.50$ on Satellite and Optdigits), and the mechanism is a simple
concentration effect: under a correct component a normal point in $d$ dimensions
sits at $\approx\!\sqrt{d}\,\sigma$ (already $\approx\!8\sigma$ at $d=64$), so
\emph{every} point exceeds $3\sigma$ and the statistic loses all resolution.
Replacing the fixed radius with the dimension-calibrated
$\tau_d=\sqrt{\chi^2_d(0.99)}$ of Eq.~\eqref{eq:taud}
\textbf{removes the saturation entirely}: window-AUC rises from $0.50$ to
$\mathbf{1.00}$ on Satellite ($d{=}36$) and from $0.50$ to $\mathbf{0.89}$ on
Optdigits ($d{=}64$), while leaving the already-solved low-dimensional cases
unchanged (Table~\ref{tab:driftcmp}, last two columns). The reference against which
this matters is MMD, which is at or near AUC $1.0$ on every dataset (lowest $0.93$
on Satellite): the calibrated unexplained-mass statistic matches or beats it on
five of the seven datasets and trails it by ${\sim}0.1$ AUC on the remaining two
(Pendigits $0.87$ vs.\ $0.98$, Optdigits $0.89$ vs.\ $1.00$)---parity rather than
dominance---while remaining self-explanatory: it reports the fraction of
the window matching no known regime. The two GMM-to-GMM divergences are weaker as
detectors and serve as ablations. The CS divergence is never the best on any
dataset and collapses to chance on the $64$-dimensional Optdigits set ($0.51$); we
retain it as a principled, closed-form baseline, not as a recommendation. The KL
surrogate is \emph{unstable}: it falls \emph{below} chance ($0.5$) on Shuttle
($0.29$) and Pendigits ($0.46$)---though it is also the strongest GMM-based
divergence on Satellite ($0.98$)---so an approximate, asymmetric KL is an erratic
rather than uniformly poor drift statistic. The mean-NLL control behaves like
a strong but uncalibrated cousin of the unexplained mass: competitive in low
and moderate dimensions (and ahead on Pendigits, $1.00$), but degrading to
$0.55$ on the $64$-dimensional Optdigits, where log-likelihoods
degenerate---the same failure the $\chi^2_d$ calibration repairs for the
per-point statistic ($0.89$ there).

\begin{table}[H]
\centering
\caption{Drift detection, window-level ROC-AUC: the two GMM divergences (CS, KL
surrogate), the mean-NLL control (generic aggregation of the GMM's point
score), the interpretable unexplained-mass GMM detector, and model-free MMD.
The interpretable GMM detector, with the fixed $3\sigma$ threshold, saturates in high
dimension; a dimension-calibrated $\chi^2_d$ threshold restores it to parity with
MMD (matching or beating it on five of seven datasets, within ${\sim}0.1$ AUC on
the rest).}
\label{tab:driftcmp}
\resizebox{\textwidth}{!}{%
\begin{tabular}{lrrrrrr}
\hline
Dataset & CS div.\ & KL surr.\ & Win.\ NLL & Unexpl.\ ($3\sigma$) & Unexpl.\ ($\chi^2_d$) & MMD \\
 & \footnotesize(GMM) & \footnotesize(GMM) & \footnotesize(GMM) & \footnotesize(interp.) & \footnotesize(interp., dim-cal.) & \footnotesize(model-free) \\
\hline
KDD-http & 0.97{\footnotesize$\pm$0.02} & 0.46{\footnotesize$\pm$0.04} & \textbf{1.00{\footnotesize$\pm$0.00}} & \textbf{1.00{\footnotesize$\pm$0.00}} & \textbf{1.00{\footnotesize$\pm$0.00}} & \textbf{1.00{\footnotesize$\pm$0.00}} \\
KDD-smtp & 0.70{\footnotesize$\pm$0.04} & 0.93{\footnotesize$\pm$0.04} & \textbf{1.00{\footnotesize$\pm$0.00}} & \textbf{1.00{\footnotesize$\pm$0.00}} & \textbf{1.00{\footnotesize$\pm$0.00}} & 0.99{\footnotesize$\pm$0.00} \\
Mammography & 0.65{\footnotesize$\pm$0.07} & 0.72{\footnotesize$\pm$0.09} & 0.99{\footnotesize$\pm$0.03} & \textbf{1.00{\footnotesize$\pm$0.00}} & \textbf{1.00{\footnotesize$\pm$0.00}} & 1.00{\footnotesize$\pm$0.00} \\
Shuttle & 0.67{\footnotesize$\pm$0.07} & 0.29{\footnotesize$\pm$0.06} & \textbf{1.00{\footnotesize$\pm$0.00}} & \textbf{1.00{\footnotesize$\pm$0.00}} & \textbf{1.00{\footnotesize$\pm$0.00}} & \textbf{1.00{\footnotesize$\pm$0.00}} \\
Pendigits & 0.56{\footnotesize$\pm$0.06} & 0.46{\footnotesize$\pm$0.05} & \textbf{1.00{\footnotesize$\pm$0.00}} & 0.87{\footnotesize$\pm$0.04} & 0.87{\footnotesize$\pm$0.20} & 0.98{\footnotesize$\pm$0.01} \\
Satellite & 0.81{\footnotesize$\pm$0.14} & 0.98{\footnotesize$\pm$0.01} & \textbf{1.00{\footnotesize$\pm$0.00}} & 0.50{\footnotesize$\pm$0.00} & \textbf{1.00{\footnotesize$\pm$0.00}} & 0.93{\footnotesize$\pm$0.09} \\
Optdigits & 0.51{\footnotesize$\pm$0.07} & 0.53{\footnotesize$\pm$0.06} & 0.55{\footnotesize$\pm$0.09} & 0.50{\footnotesize$\pm$0.00} & 0.89{\footnotesize$\pm$0.12} & \textbf{1.00{\footnotesize$\pm$0.00}} \\
\hline
\multicolumn{6}{l}{\footnotesize Window-level ROC-AUC (clean vs.\ 10\% contaminated), mean\,$\pm$\,std over five seeds;}\\
\multicolumn{6}{l}{\footnotesize best per row in bold. Chance $=0.5$. `Unexpl.'\ is the interpretable window fraction}\\
\multicolumn{6}{l}{\footnotesize beyond a threshold: fixed $3\sigma$, which saturates in high dimension, vs.\ the}\\
\multicolumn{6}{l}{\footnotesize dimension-calibrated $\chi^2_d(0.99)$ quantile, which restores it.}\\
\end{tabular}}

\end{table}

\paragraph{Against the recent model-free field.}
Table~\ref{tab:driftbase} reports the same protocol for the seven literature
baselines. The picture that emerges is one of parity without dominance: no
detector, including MMD, is best everywhere, and every baseline has at least one
benchmark where it falters---C2ST drops to $0.88$ on Satellite, LSDD to $0.79$
on KDD-smtp, sliced Wasserstein to $0.89$ on Shuttle, QuantTree to $0.83$ on
Optdigits, EI-kMeans to $0.50$--$0.67$ where its validity rules force a
two-cluster partition (Optdigits, KDD-http), and SPLL collapses to
$0.41$ on the $64$-dimensional Optdigits, the same high-dimensional regime that
breaks the fixed-$3\sigma$ unexplained mass (its pooled covariance becomes
near-singular, so the Mahalanobis geometry degenerates). The calibrated
unexplained-mass statistic is best or tied-best on five of seven datasets, and
its worst case ($0.87$, Pendigits) sits between LSDD's ($0.79$) and C2ST's or
sliced Wasserstein's ($0.88$, $0.89$), though the strongest worst cases belong
to PCA-CD ($0.92$) and MMD ($0.93$). The repair is also not free on Pendigits:
the calibrated statistic keeps the fixed-$\tau$ mean there ($0.87$) but its
seed variance grows ($\pm0.04\to\pm0.20$).
The point of the comparison is not that our statistic wins---it is
that \emph{nothing here is bought with accuracy}: the only detector in the table
whose score is a directly auditable quantity (the fraction of the window
explained by no known regime, traceable to named regimes and named points)
performs on par with the modern model-free field.

\begin{table}[H]
\centering
\caption{The dimension-calibrated unexplained mass against seven recent
model-free drift baselines: window-level ROC-AUC (clean vs.\ 10\%
contaminated), mean\,$\pm$\,std over five seeds, best per row in bold (same
protocol as Table~\ref{tab:driftcmp}). C2ST: domain-classifier two-sample
test; SPLL: semi-parametric log-likelihood; PCA-CD: KS statistic along
reference principal components; LSDD: least-squares density difference; SW:
sliced Wasserstein-1 distance; QTree: QuantTree Pearson statistic; EI-kM:
EI-kMeans $\chi^2$ score. No method dominates; the interpretable
statistic is best or tied-best on five of seven datasets, and only it yields a
regime-level explanation of each alarm.}
\label{tab:driftbase}
\resizebox{\textwidth}{!}{%
\begin{tabular}{lrrrrrrrrr}
\hline
Dataset & Unexpl.\ ($\chi^2_d$) & MMD & C2ST & SPLL & PCA-CD & LSDD & SW dist. & QTree & EI-kM \\
\hline
KDD-http & \textbf{1.00{\footnotesize$\pm$0.00}} & \textbf{1.00{\footnotesize$\pm$0.00}} & 0.99{\footnotesize$\pm$0.02} & 0.71{\footnotesize$\pm$0.14} & 0.99{\footnotesize$\pm$0.01} & 0.90{\footnotesize$\pm$0.06} & \textbf{1.00{\footnotesize$\pm$0.00}} & \textbf{1.00{\footnotesize$\pm$0.00}} & 0.67{\footnotesize$\pm$0.20} \\
KDD-smtp & \textbf{1.00{\footnotesize$\pm$0.00}} & 0.99{\footnotesize$\pm$0.00} & 0.93{\footnotesize$\pm$0.03} & \textbf{1.00{\footnotesize$\pm$0.00}} & 0.97{\footnotesize$\pm$0.01} & 0.79{\footnotesize$\pm$0.08} & \textbf{1.00{\footnotesize$\pm$0.00}} & \textbf{1.00{\footnotesize$\pm$0.00}} & 0.90{\footnotesize$\pm$0.19} \\
Mammography & \textbf{1.00{\footnotesize$\pm$0.00}} & \textbf{1.00{\footnotesize$\pm$0.00}} & 0.94{\footnotesize$\pm$0.05} & 0.98{\footnotesize$\pm$0.02} & 0.95{\footnotesize$\pm$0.04} & 0.98{\footnotesize$\pm$0.02} & \textbf{1.00{\footnotesize$\pm$0.00}} & 0.97{\footnotesize$\pm$0.03} & \textbf{1.00{\footnotesize$\pm$0.00}} \\
Shuttle & \textbf{1.00{\footnotesize$\pm$0.00}} & \textbf{1.00{\footnotesize$\pm$0.00}} & 0.98{\footnotesize$\pm$0.02} & 0.78{\footnotesize$\pm$0.05} & 0.99{\footnotesize$\pm$0.00} & \textbf{1.00{\footnotesize$\pm$0.00}} & 0.89{\footnotesize$\pm$0.03} & \textbf{1.00{\footnotesize$\pm$0.00}} & 0.88{\footnotesize$\pm$0.19} \\
Pendigits & 0.87{\footnotesize$\pm$0.20} & 0.98{\footnotesize$\pm$0.01} & 0.91{\footnotesize$\pm$0.07} & \textbf{1.00{\footnotesize$\pm$0.00}} & 0.96{\footnotesize$\pm$0.03} & 0.95{\footnotesize$\pm$0.03} & 0.98{\footnotesize$\pm$0.01} & 0.96{\footnotesize$\pm$0.05} & 0.94{\footnotesize$\pm$0.07} \\
Satellite & \textbf{1.00{\footnotesize$\pm$0.00}} & 0.93{\footnotesize$\pm$0.09} & 0.88{\footnotesize$\pm$0.06} & \textbf{1.00{\footnotesize$\pm$0.00}} & 0.92{\footnotesize$\pm$0.07} & 0.85{\footnotesize$\pm$0.17} & \textbf{1.00{\footnotesize$\pm$0.00}} & 0.86{\footnotesize$\pm$0.09} & \textbf{1.00{\footnotesize$\pm$0.01}} \\
Optdigits & 0.89{\footnotesize$\pm$0.12} & \textbf{1.00{\footnotesize$\pm$0.00}} & 0.98{\footnotesize$\pm$0.02} & 0.41{\footnotesize$\pm$0.12} & 0.98{\footnotesize$\pm$0.01} & \textbf{1.00{\footnotesize$\pm$0.00}} & 0.94{\footnotesize$\pm$0.08} & 0.83{\footnotesize$\pm$0.09} & 0.50{\footnotesize$\pm$0.13} \\
\hline
\end{tabular}}

\end{table}

\begin{figure}[H]
\centering
\includegraphics[width=\textwidth]{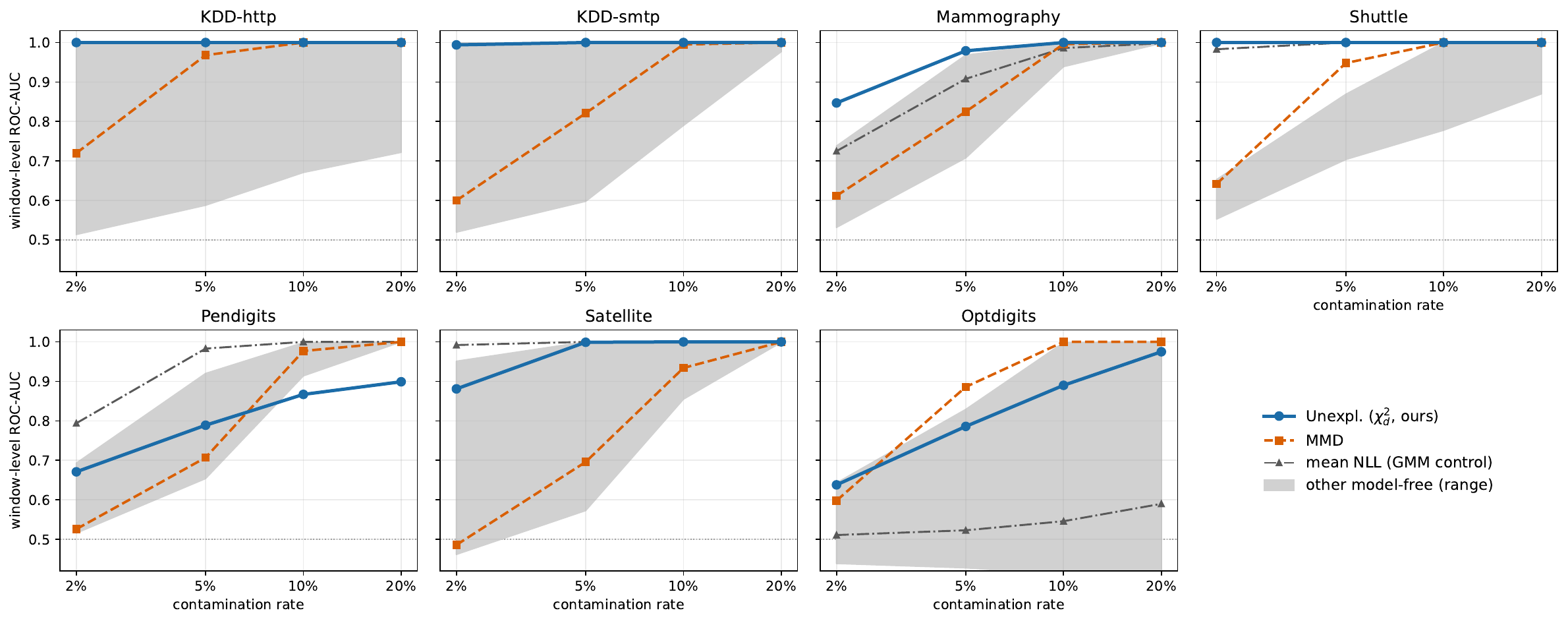}
\caption{Window-level ROC-AUC vs.\ contamination rate ($2$--$20\%$, log scale;
five seeds). The calibrated unexplained mass (ours), MMD, and the mean-NLL
control are drawn individually; the shaded band is the min--max range of the
other seven model-free detectors. Sample-level detectors converge to AUC
${\approx}1$ at heavy contamination but fade toward chance as contamination
becomes sparse; the model-based statistics stay strong, with the calibrated
fraction the most robust (note the control's collapse on Optdigits).}
\label{fig:sweep}
\end{figure}

\paragraph{Sensitivity to the contamination rate.}
The $10\%$ contamination of Tables~\ref{tab:driftcmp}--\ref{tab:driftbase} is
one point on a spectrum; drift in the wild can be far subtler. Repeating the
experiment at $2\%$, $5\%$ and $20\%$ contamination
(Figure~\ref{fig:sweep}) separates the detectors into two regimes. At $20\%$
every sample-level detector is at or near AUC $1.0$ on every dataset (only
SPLL and EI-kMeans retain their high-dimensional weaknesses from
Table~\ref{tab:driftbase}). As the contamination becomes sparse, however, the \emph{sample-level}
detectors fade together toward chance---at $2\%$ (ten anomalous points in a
$500$-point window) MMD averages $0.60$ across the seven benchmarks, C2ST
$0.60$, PCA-CD $0.57$, LSDD $0.54$, and the partition detectors $0.55$--$0.58$,
because ten points barely move a mean embedding, a classifier boundary, or a
histogram. The two \emph{model-based, per-point} statistics degrade far more
gracefully: the calibrated unexplained mass averages $0.86$ (best or tied-best
on four of seven datasets, e.g.\ $1.00$ vs.\ MMD's $0.72$ on KDD-http and
$1.00$ vs.\ $0.64$ on Shuttle) and SPLL $0.71$, because a
handful of points landing outside every regime moves a per-point fraction
decisively even when sample-level distances cannot see them. Sliced
Wasserstein is the strongest model-free detector in this regime (mean $0.73$)
yet still trails clearly. (``Most gracefully'' here means concretely: the
highest mean AUC at every contamination rate below $10\%$---$0.94$ vs.\ at
most $0.87$ at $5\%$, $0.86$ vs.\ at most $0.73$ at $2\%$---not a win on
every dataset; SPLL is stronger on Pendigits and Satellite at $2\%$.) This
inverts the usual framing: the interpretable
statistic is not purchasing explanation with accuracy---in the evaluated
sparse-contamination setting, it has the highest mean accuracy of all
detectors tested. An important control disentangles what earns this: the
mean-NLL aggregation of the same GMM's point score matches the unexplained
mass at $2\%$ (mean $0.86$)---so the sparse-regime advantage comes from being
\emph{model-based and per-point}, not from the unexplained-mass construction
alone. What the calibrated fraction adds over the generic aggregation is
robustness and semantics: mean NLL collapses at $d{=}64$ under $10\%$
contamination ($0.55$ vs.\ our $0.89$), its one-sided value inverts under
in-support re-weighting (Table~\ref{tab:reweight}), and its numeric value,
unlike a fraction of unexplained points, explains nothing.

\subsection{Two kinds of drift: contamination vs.\ in-support re-weighting}
\label{InSupport}
The experiment above injects point-outliers, so a window drifts by acquiring
anomalies that lie \emph{outside} the normal support---\emph{out-of-support}
(novel-regime) drift. But we motivated drift more
broadly (Section~\ref{Introduction}) as the case where individual points may each
look normal while their \emph{collective} distribution has changed. These are
genuinely different, and they need different detectors. To isolate the second
case, we build drifted windows that contain \emph{only real normal points} but
drawn with \emph{re-weighted regime proportions} (we assign each normal point to
its nearest reference component and over-sample the dominant regimes). No point is
an outlier---the ``outlier fraction'' column of Table~\ref{tab:reweight} stays at
the clean baseline---yet the distribution has genuinely shifted.

The result is a clean dissociation. On the datasets where the re-weighting truly
keeps every point in support (KDD-http, KDD-smtp and Mammography, outlier fraction
$\leq2\%$), the interpretable \emph{unexplained-mass} statistic is \emph{blind}:
its AUC is at or below chance ($0.47$, $0.17$, $0.01$), because clean and drifted
windows contain the same in-support points. The closed-form CS divergence is
unreliable ($0.26$--$0.74$ across datasets, strong on some, below chance on
others). The distribution-level detectors, by contrast, see the
re-weighting: MMD (AUC $0.98$--$1.0$ on all seven), LSDD ($0.99$--$1.0$
throughout), PCA-CD ($\geq0.88$ everywhere), QuantTree ($\geq0.89$
everywhere), and C2ST on all but KDD-http
($0.76$ there, $\geq0.99$ elsewhere); sliced Wasserstein is strong on five of
seven but drops to $0.66$ on Shuttle and $0.84$ on KDD-http, and EI-kMeans
detects on six of seven ($0.90$--$1.0$) but falls to $0.62$ on KDD-http. SPLL is
instructive in a different way: re-weighting toward dominant regimes
\emph{concentrates} the window near cluster centres, so its
reference-to-window direction moves \emph{down} rather than up, and only the
opposite direction---scoring the reference sample under the window's
clusters, included in Kuncheva's max-of-both-directions
definition---catches the shift. The two-directional statistic detects the
re-weighting on six of seven datasets ($0.79$--$1.0$, middling only on
Optdigits at $0.79$) but stays weak on Shuttle ($0.30$): a one-sided
mean-Mahalanobis monitor would miss in-support drift entirely, and even the
two-sided form is the least reliable of the distribution-level detectors
here. The mean-NLL control shows the one-sided failure in its purest form:
concentrating a window on dominant regimes \emph{raises} its likelihood, so
the statistic moves the wrong way and its AUC inverts to ${\approx}0$ on four
of seven datasets. (On the higher-dimensional sets
Satellite and Optdigits the
outlier fraction is $1.0$ not because re-weighting creates outliers but because of
the same fixed-$\tau=3\sigma$ saturation as Table~\ref{tab:driftcmp}---every normal
point sits beyond $3\sigma$---so the unexplained-mass AUC there is pinned at exactly
$0.50$; the three low-dimensional datasets, with outlier fraction $\leq2\%$,
isolate the in-support effect cleanly.) This dissociation is structural, not
incidental---clean and drifted windows contain the same in-support points, so no
statistic built on per-point regime membership can separate them---and it defines
the division of labour: the unexplained-mass statistic detects \emph{novel-regime}
drift (contamination, new operating modes) \emph{and explains it}; detecting
\emph{in-support} drift (a shift in the mix of known regimes) requires a
distribution-level test such as MMD, which detects it robustly but supplies no
explanation of its alarms.

\begin{table}[H]
\centering
\caption{In-support drift (regime re-weighting): window-level ROC-AUC, mean
over five seeds (std $\leq0.27$; full values in \texttt{results.json}), best
per row in bold. On the clean in-support cases
(top three rows) the interpretable statistic is blind, and the distribution-level
detectors (MMD, C2ST, PCA-CD, LSDD, SW, QTree, EI-kM) detect the shift; SPLL
needs its window-to-reference direction to see it, and the one-sided mean-NLL
control inverts outright. `Outl.\ frac.'\ is the share of drifted-window points beyond
$3\sigma$ of every regime: the top three rows are clean in-support cases
(frac.\ $\leq2\%$); on the rest the fraction is dominated by the fixed-$\tau$
saturation of Table~\ref{tab:driftcmp}, not by re-weighting. Contrast with
Table~\ref{tab:driftbase} (contamination), where
the interpretable statistic matches the same field.}
\label{tab:reweight}
\resizebox{\textwidth}{!}{%
\begin{tabular}{lrrrrrrrrrrrr}
\hline
Dataset & CS & Unexpl.\ & NLL & MMD & C2ST & SPLL & PCA-CD & LSDD & SW & QTree & EI-kM & Outl.\ frac.\ \\
\hline
KDD-http & 0.74 & 0.47 & 0.00 & 0.98 & 0.76 & \textbf{1.00} & 0.88 & 0.99 & 0.83 & 0.99 & 0.62 & 0.000 \\
KDD-smtp & 0.74 & 0.17 & 0.00 & \textbf{1.00} & \textbf{1.00} & 0.98 & \textbf{1.00} & \textbf{1.00} & \textbf{1.00} & \textbf{1.00} & \textbf{1.00} & 0.001 \\
Mammography & 0.43 & 0.01 & 0.00 & \textbf{1.00} & \textbf{1.00} & \textbf{1.00} & \textbf{1.00} & \textbf{1.00} & \textbf{1.00} & \textbf{1.00} & \textbf{1.00} & 0.018 \\
Shuttle & 0.38 & 0.64 & 0.32 & \textbf{1.00} & 0.99 & 0.30 & \textbf{1.00} & \textbf{1.00} & 0.66 & \textbf{1.00} & 0.90 & 0.188 \\
Pendigits & 0.63 & 0.41 & 0.91 & \textbf{1.00} & 0.99 & 0.92 & 0.97 & \textbf{1.00} & \textbf{1.00} & 0.96 & \textbf{1.00} & 0.690 \\
Satellite & 0.56 & 0.50 & 0.00 & \textbf{1.00} & \textbf{1.00} & \textbf{1.00} & \textbf{1.00} & 0.99 & \textbf{1.00} & 0.89 & \textbf{1.00} & 1.000 \\
Optdigits & 0.26 & 0.50 & 0.04 & \textbf{1.00} & \textbf{1.00} & 0.79 & 0.91 & \textbf{1.00} & 0.89 & 0.94 & 0.95 & 1.000 \\
\hline
\end{tabular}}

\end{table}

\subsection{Interpretability: attributing alarms to regimes}\label{Interpret}
Unlike a black-box detector, the GMM lets every alarm be \emph{explained} in
terms of its named components, or ``regimes''. We illustrate this on KDD-http,
whose BIC-selected model has nine components---five carrying the bulk of the
mass---with the weights shown in Table~\ref{tab:regimes}. We use two
interpretable quantities.

\begin{figure}[H]
\centering
\includegraphics[width=\textwidth]{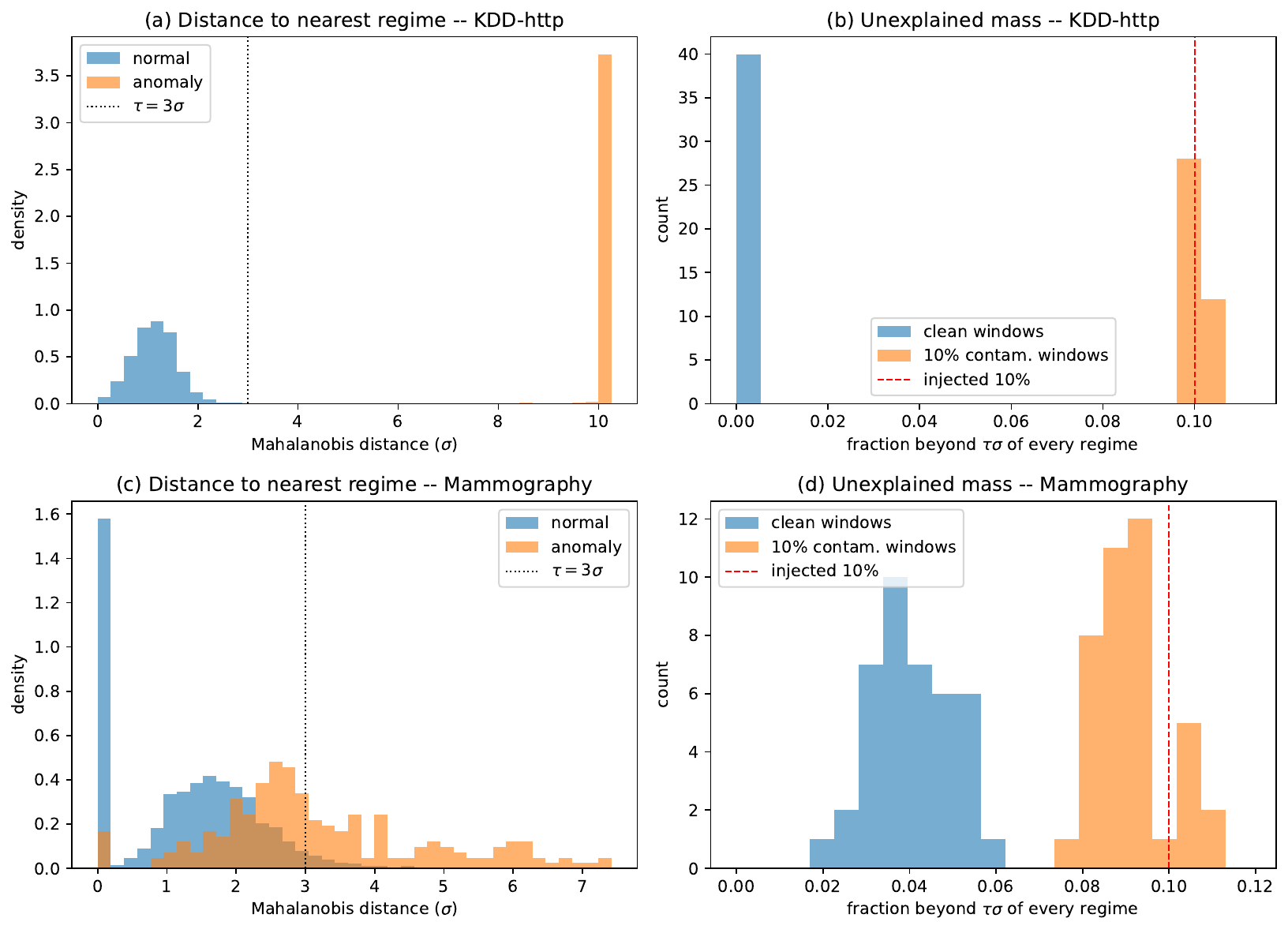}
\caption{Interpretability on KDD-http (top) and Mammography (bottom). (a,c)
Mahalanobis distance to the nearest normal regime: normal points cluster within a
few $\sigma$, anomalies lie far outside (on Mammography the anomaly distances are
genuinely spread, not a single spike). (b,d) Unexplained mass per window (fraction
beyond $\tau\sigma$ of every regime): clean vs.\ $10\%$-contaminated windows. The
recovery of the injected $10\%$ is sharp on KDD-http and looser on Mammography
($4\%$ vs.\ $9\%$).}
\label{fig:interpret}
\end{figure}

\emph{Regime attribution (point level).} For an observation $x$ we report the
nearest component $k^\star(x)=\arg\min_k \Delta_k(x)$, where
$\Delta_k(x)=\sqrt{(x-\mu_k)^\top\Sigma_k^{-1}(x-\mu_k)}$ is the Mahalanobis
distance, together with $\Delta_{k^\star}(x)$ itself---\emph{how far outside} the
nearest known regime the point lies. On KDD-http, normal test points sit a median
of $1.1\sigma$ (95th percentile $1.8\sigma$) from their nearest regime, whereas
flagged anomalies sit a median of $10.3\sigma$ away
(Figure~\ref{fig:interpret}a). An alarm therefore comes with a concrete
sentence: \emph{``this record's nearest normal regime is $k^\star$, but it lies
$10\sigma$ outside it.''} We caution that on KDD-http the anomaly distances are
tightly concentrated (median $\approx$ 95th percentile), an artefact of duplicated
attack records. On Mammography the same separation holds without that artefact:
anomaly distances are genuinely spread (median $3.1\sigma$, 95th percentile
$14\sigma$) yet clearly separated from normal points (median $1.4\sigma$),
confirming the effect is not specific to one dataset.

\emph{Unexplained mass (window level).} For drift we report the fraction of a
window whose nearest regime is still farther than $\tau=3\sigma$---the share of
the window matching \emph{no} known regime. Clean windows have a mean unexplained
mass of $0.1\%$, whereas $10\%$-contaminated windows have $10.1\%$, recovering the
injected contamination almost exactly (Figure~\ref{fig:interpret}b). This turns a
drift alarm into an actionable statement: \emph{``$\sim\!10\%$ of this window
matches no known operating regime.''}

\begin{table}[H]
\centering
\caption{Interpretable regimes (mixture components) for KDD-http.}
\label{tab:regimes}
\begin{tabular}{lrrrrrrrrr}
\hline
Regime (by weight) & 1 & 2 & 3 & 4 & 5 & 6 & 7 & 8 & 9 \\
Weight $\pi_k$ (\%) & 28.8 & 22.0 & 19.7 & 19.4 & 9.3 & 0.4 & 0.3 & 0.1 & 0.0 \\
\hline
\multicolumn{10}{l}{\footnotesize KDD-http: 9 components, of which 5 carry $\geq\!1\%$ of the mass. Normal points lie a median 1.1$\sigma$}\\
\multicolumn{10}{l}{\footnotesize (p95 1.8$\sigma$) from their nearest regime, whereas flagged anomalies lie a median 10.3$\sigma$ away.}\\
\end{tabular}

\end{table}

\section{Discussion}\label{Discussion}
\paragraph{Interpretability.} The GMM's advantage over black-box detectors is
not raw accuracy but explanation: each alarm can be attributed to the mixture
component (regime) whose responsibility dropped or whose region the data
vacated, together with the divergence magnitude. This matters in condition
monitoring, where operators need a reason, not just a flag.

\paragraph{Dimensionality.} Density estimation degrades as $d$ grows; the GMM
remains competitive up to $d{=}36$ (Satellite) here, but for higher dimensions we
recommend a dimensionality-reduction front-end (PCA or an autoencoder) before
the mixture, or a switch to a different detector family.

\paragraph{When to use what.} Use likelihood scoring~\eqref{eq:nll} for point
anomalies and the $\chi^2_d$-calibrated unexplained-mass
statistic~\eqref{eq:unexplained} for drift, falling back to a model-free kernel
two-sample test such as MMD \citep{Gretton2012} when in-support re-weighting must
also be caught. The GMM route earns its place when interpretable regimes are
wanted or the mixture assumption is reasonable; the closed-form CS
divergence~\eqref{eq:csgmm} is empirically dominated and serves as an ablation.

\paragraph{Limitations.} The contamination experiments inject labelled
point-anomalies from each benchmark's anomaly class, so the drift they create
is by construction \emph{out-of-support}; the in-support experiment
(Section~\ref{InSupport}) probes the complementary case, but drifts that mix
both characters, and drifts in the tail shape of existing regimes, are not
covered. A fuller study would also add more high-dimensional benchmarks. The
drift experiments assume windows large enough to fit a GMM, trading latency
for the ability to see distributional change, and the ``regimes'' that make
alarms interpretable are model-defined operational concepts---components of a
BIC-selected fit---not necessarily physically distinct modes of the system.
The interpretability case study is shown on one dataset for clarity, though
the same attribution applies to any GMM.

\section{Conclusion}\label{Conclusion}
The unexplained-mass statistic---the fraction of a stream window matching no known
regime of a GMM fitted to normal data---is a drift signal that explains itself, and
we showed that its well-known high-dimensional failure is a simple concentration
effect with a complete repair: calibrating the regime radius to the data dimension
via the $\chi^2_d(0.99)$ quantile lifts window-AUC from chance to $1.00$
(Satellite, $d{=}36$) and $0.89$ (Optdigits, $d{=}64$), placing the interpretable
detector broadly competitive with eight model-free detectors from the recent
drift literature---MMD, a classifier two-sample test, PCA-based detection,
LSDD, sliced Wasserstein, SPLL, QuantTree and EI-kMeans---across the tested
dimensions: best or tied-best on five of seven benchmarks at $10\%$
contamination (clearly beaten on Pendigits and Optdigits), and with the
highest mean accuracy of all detectors tested as contamination becomes
sparse. The statistic's
scope is equally sharp: it detects and explains \emph{novel-regime} drift, and is
blind by construction to \emph{in-support} re-weighting of known regimes, where MMD
detects robustly but explains nothing---a division of labour, not a horse race. The
closed-form Cauchy--Schwarz divergence is empirically dominated and retained as an
ablation. Supporting all of this, the same BIC-selected, EVT-thresholded density
model is a competent point-anomaly detector on par with standard baselines, and
every alarm it raises is attributable: which regime the data matches or fails to
match, and by how much. Future work will add streaming/online GMM updates, deep
baselines, and a dimensionality-reduction front-end for high-dimensional data.

\paragraph{Reproducibility.} All datasets are public; the code, dataset loaders,
divergence implementations and experiment scripts are released in the
\texttt{prototype/} directory accompanying this paper and provided as
supplementary material with this submission. All randomness is seeded;
package versions are pinned in \texttt{requirements.txt}. (Terminology: the
point-anomaly benchmark uses five random \emph{train/test splits}; the drift
experiments use five \emph{seeds}, each fixing the split, the reference
sample, and all window draws.) The drift comparison
(Tables~\ref{tab:driftcmp}--\ref{tab:driftbase}) uses \emph{five random seeds}, each contributing $40$
clean and $40$ contaminated windows per dataset; the reported values are the mean and
the $\pm$ columns the standard deviation over the five seeds. Orderings within
$\sim\!0.05$ AUC should be read as ties.

\bibliographystyle{plainnat}
\bibliography{references}

\begin{thebibliography}{59}
\providecommand{\natexlab}[1]{#1}
\providecommand{\url}[1]{\texttt{#1}}
\expandafter\ifx\csname urlstyle\endcsname\relax
  \providecommand{\doi}[1]{doi: #1}\else
  \providecommand{\doi}{doi: \begingroup \urlstyle{rm}\Url}\fi

\bibitem[Assis and de~Souza(2025)]{assis2025adwinu}
Daniel~Nowak Assis and Vin{\'i}cius M.~A. de~Souza.
\newblock {ADWIN-U}: adaptive windowing for unsupervised drift detection on
  data streams.
\newblock \emph{Knowledge and Information Systems}, 67\penalty0 (11):\penalty0
  10005--10034, 2025.
\newblock \doi{10.1007/s10115-025-02523-1}.

\bibitem[Biernacki et~al.(2000)Biernacki, Celeux, and Govaert]{Biernacki2000}
Christophe Biernacki, Gilles Celeux, and G{\'e}rard Govaert.
\newblock Assessing a mixture model for clustering with the integrated
  completed likelihood.
\newblock \emph{IEEE Transactions on Pattern Analysis and Machine
  Intelligence}, 22\penalty0 (7):\penalty0 719--725, 2000.

\bibitem[Bishop(2006)]{Bishop2006}
Christopher~M. Bishop.
\newblock \emph{Pattern Recognition and Machine Learning}.
\newblock Springer, 2006.

\bibitem[Boracchi et~al.(2018)Boracchi, Carrera, Cervellera, and
  Macci{\`o}]{boracchi2018quanttree}
Giacomo Boracchi, Diego Carrera, Cristiano Cervellera, and Danilo Macci{\`o}.
\newblock {QuantTree}: Histograms for change detection in multivariate data
  streams.
\newblock In \emph{Proceedings of the 35th International Conference on Machine
  Learning (ICML)}, volume~80 of \emph{Proceedings of Machine Learning
  Research}, pages 639--648, 2018.

\bibitem[Breunig et~al.(2000)Breunig, Kriegel, Ng, and Sander]{Breunig2000}
Markus~M. Breunig, Hans-Peter Kriegel, Raymond~T. Ng, and J{\"o}rg Sander.
\newblock {LOF}: Identifying density-based local outliers.
\newblock In \emph{ACM SIGMOD International Conference on Management of Data},
  pages 93--104, 2000.

\bibitem[Bu et~al.(2018)Bu, Alippi, and Zhao]{bu2018lsdd}
Li~Bu, Cesare Alippi, and Dongbin Zhao.
\newblock A pdf-free change detection test based on density difference
  estimation.
\newblock \emph{IEEE Transactions on Neural Networks and Learning Systems},
  29\penalty0 (2):\penalty0 324--334, 2018.
\newblock \doi{10.1109/TNNLS.2016.2619909}.

\bibitem[Cerqueira et~al.(2023)Cerqueira, Gomes, Bifet, and
  Torgo]{cerqueira2023studd}
V{\'i}tor Cerqueira, Heitor~Murilo Gomes, Albert Bifet, and Lu{\'i}s Torgo.
\newblock {STUDD}: a student--teacher method for unsupervised concept drift
  detection.
\newblock \emph{Machine Learning}, 112\penalty0 (11):\penalty0 4351--4378,
  2023.
\newblock \doi{10.1007/s10994-022-06188-7}.

\bibitem[Chandola et~al.(2009)Chandola, Banerjee, and Kumar]{Chandola2009}
Varun Chandola, Arindam Banerjee, and Vipin Kumar.
\newblock Anomaly detection: A survey.
\newblock \emph{ACM Computing Surveys}, 41\penalty0 (3):\penalty0 1--58, 2009.

\bibitem[Coles(2001)]{Coles2001}
Stuart Coles.
\newblock \emph{An Introduction to Statistical Modeling of Extreme Values}.
\newblock Springer, 2001.

\bibitem[de~Souza et~al.(2021)de~Souza, Parmezan, Chowdhury, and
  Mueen]{desouza2021ibdd}
Vin{\'i}cius M.~A. de~Souza, Antonio R.~S. Parmezan, Farhan~A. Chowdhury, and
  Abdullah Mueen.
\newblock Efficient unsupervised drift detector for fast and high-dimensional
  data streams.
\newblock \emph{Knowledge and Information Systems}, 63\penalty0 (6):\penalty0
  1497--1527, 2021.
\newblock \doi{10.1007/s10115-021-01564-6}.

\bibitem[Dempster et~al.(1977)Dempster, Laird, and Rubin]{Dempster1977}
A.~P. Dempster, N.~M. Laird, and D.~B. Rubin.
\newblock Maximum likelihood from incomplete data via the {EM} algorithm.
\newblock \emph{Journal of the Royal Statistical Society, Series B},
  39\penalty0 (1):\penalty0 1--38, 1977.

\bibitem[Frittoli et~al.(2023)Frittoli, Carrera, and
  Boracchi]{frittoli2023quanttree}
Luca Frittoli, Diego Carrera, and Giacomo Boracchi.
\newblock Nonparametric and online change detection in multivariate datastreams
  using {QuantTree}.
\newblock \emph{IEEE Transactions on Knowledge and Data Engineering},
  35\penalty0 (8), 2023.
\newblock \doi{10.1109/TKDE.2022.3202730}.

\bibitem[Fukunaga(1990)]{Fukunaga1990}
Keinosuke Fukunaga.
\newblock \emph{Introduction to Statistical Pattern Recognition}.
\newblock Academic Press, 2nd edition, 1990.

\bibitem[Gama et~al.(2014)Gama, {\v{Z}}liobait{\.e}, Bifet, Pechenizkiy, and
  Bouchachia]{Gama2014}
Jo{\~a}o Gama, Indr{\.e} {\v{Z}}liobait{\.e}, Albert Bifet, Mykola Pechenizkiy,
  and Abdelhamid Bouchachia.
\newblock A survey on concept drift adaptation.
\newblock \emph{ACM Computing Surveys}, 46\penalty0 (4):\penalty0 1--37, 2014.

\bibitem[Gemaque et~al.(2020)Gemaque, Costa, Giusti, and dos
  Santos]{gemaque2020overview}
Rosana~Noronha Gemaque, Albert Fran{\c{c}}a~Josu{\'a} Costa, Rafael Giusti, and
  Eulanda~Miranda dos Santos.
\newblock An overview of unsupervised drift detection methods.
\newblock \emph{WIREs Data Mining and Knowledge Discovery}, 10\penalty0
  (6):\penalty0 e1381, 2020.
\newblock \doi{10.1002/widm.1381}.

\bibitem[Goldberger et~al.(2003)Goldberger, Gordon, and
  Greenspan]{Goldberger2003}
Jacob Goldberger, Shiri Gordon, and Hayit Greenspan.
\newblock An efficient image similarity measure based on approximations of
  {KL}-divergence between two {G}aussian mixtures.
\newblock In \emph{IEEE International Conference on Computer Vision (ICCV)},
  pages 487--493, 2003.

\bibitem[Goldenberg and Webb(2019)]{goldenberg2019survey}
Igor Goldenberg and Geoffrey~I. Webb.
\newblock Survey of distance measures for quantifying concept drift and shift
  in numeric data.
\newblock \emph{Knowledge and Information Systems}, 60\penalty0 (2):\penalty0
  591--615, 2019.
\newblock \doi{10.1007/s10115-018-1257-z}.

\bibitem[Goldenberg and Webb(2020)]{goldenberg2020pca}
Igor Goldenberg and Geoffrey~I. Webb.
\newblock {PCA}-based drift and shift quantification framework for
  multidimensional data.
\newblock \emph{Knowledge and Information Systems}, 62\penalty0 (7):\penalty0
  2835--2854, 2020.
\newblock \doi{10.1007/s10115-020-01438-3}.

\bibitem[G{\"o}z{\"u}a{\c{c}}{\i}k et~al.(2019)G{\"o}z{\"u}a{\c{c}}{\i}k,
  B{\"u}y{\"u}k{\c{c}}ak{\i}r, Bonab, and Can]{gozuacik2019d3}
{\"O}mer G{\"o}z{\"u}a{\c{c}}{\i}k, Alican B{\"u}y{\"u}k{\c{c}}ak{\i}r, Hamed
  Bonab, and Fazli Can.
\newblock Unsupervised concept drift detection with a discriminative
  classifier.
\newblock In \emph{Proceedings of the 28th ACM International Conference on
  Information and Knowledge Management (CIKM)}, pages 2365--2368, 2019.
\newblock \doi{10.1145/3357384.3358144}.

\bibitem[Greco et~al.(2025)Greco, Vacchetti, Apiletti, and
  Cerquitelli]{greco2025driftlens}
Salvatore Greco, Bartolomeo Vacchetti, Daniele Apiletti, and Tania Cerquitelli.
\newblock Unsupervised concept drift detection from deep learning
  representations in real-time.
\newblock \emph{IEEE Transactions on Knowledge and Data Engineering},
  37\penalty0 (10):\penalty0 6232--6245, 2025.
\newblock \doi{10.1109/TKDE.2025.3593123}.

\bibitem[Gretton et~al.(2012)Gretton, Borgwardt, Rasch, Sch{\"o}lkopf, and
  Smola]{Gretton2012}
Arthur Gretton, Karsten~M. Borgwardt, Malte~J. Rasch, Bernhard Sch{\"o}lkopf,
  and Alexander Smola.
\newblock A kernel two-sample test.
\newblock \emph{Journal of Machine Learning Research}, 13:\penalty0 723--773,
  2012.

\bibitem[Halstead et~al.(2023)Halstead, Koh, Riddle, Pechenizkiy, and
  Bifet]{halstead2023combining}
Ben Halstead, Yun~Sing Koh, Patricia Riddle, Mykola Pechenizkiy, and Albert
  Bifet.
\newblock Combining diverse meta-features to accurately identify recurring
  concept drift in data streams.
\newblock \emph{ACM Transactions on Knowledge Discovery from Data}, 17\penalty0
  (8):\penalty0 107:1--107:36, 2023.
\newblock \doi{10.1145/3587098}.

\bibitem[Han et~al.(2022)Han, Hu, Huang, Jiang, and Zhao]{han2022adbench}
Songqiao Han, Xiyang Hu, Hailiang Huang, Minqi Jiang, and Yue Zhao.
\newblock {ADBench}: Anomaly detection benchmark.
\newblock In \emph{Advances in Neural Information Processing Systems 35
  (NeurIPS), Datasets and Benchmarks Track}, 2022.

\bibitem[Haug et~al.(2022)Haug, Braun, Z{\"u}rn, and Kasneci]{haug2022cdleeds}
Johannes Haug, Alexander Braun, Stefan Z{\"u}rn, and Gjergji Kasneci.
\newblock Change detection for local explainability in evolving data streams.
\newblock In \emph{Proceedings of the 31st ACM International Conference on
  Information \& Knowledge Management (CIKM)}, pages 706--716, 2022.
\newblock \doi{10.1145/3511808.3557257}.

\bibitem[Hershey and Olsen(2007)]{Hershey2007}
John~R. Hershey and Peder~A. Olsen.
\newblock Approximating the {K}ullback-{L}eibler divergence between {G}aussian
  mixture models.
\newblock In \emph{IEEE International Conference on Acoustics, Speech and
  Signal Processing (ICASSP)}, pages IV--317--IV--320, 2007.

\bibitem[Hinder et~al.(2020)Hinder, Artelt, and Hammer]{hinder2020dawidd}
Fabian Hinder, Andr{\'e} Artelt, and Barbara Hammer.
\newblock Towards non-parametric drift detection via dynamic adapting window
  independence drift detection ({DAWIDD}).
\newblock In \emph{Proceedings of the 37th International Conference on Machine
  Learning (ICML)}, volume 119 of \emph{Proceedings of Machine Learning
  Research}, pages 4249--4259, 2020.

\bibitem[Hinder et~al.(2023)Hinder, Vaquet, Brinkrolf, and
  Hammer]{hinder2023modelbased}
Fabian Hinder, Valerie Vaquet, Johannes Brinkrolf, and Barbara Hammer.
\newblock Model-based explanations of concept drift.
\newblock \emph{Neurocomputing}, 555:\penalty0 126640, 2023.
\newblock \doi{10.1016/j.neucom.2023.126640}.

\bibitem[Hinder et~al.(2024{\natexlab{a}})Hinder, Vaquet, and
  Hammer]{hinder2024onePartA}
Fabian Hinder, Valerie Vaquet, and Barbara Hammer.
\newblock One or two things we know about concept drift---a survey on
  monitoring in evolving environments. {Part A}: detecting concept drift.
\newblock \emph{Frontiers in Artificial Intelligence}, 7:\penalty0 1330257,
  2024{\natexlab{a}}.
\newblock \doi{10.3389/frai.2024.1330257}.

\bibitem[Hinder et~al.(2024{\natexlab{b}})Hinder, Vaquet, and
  Hammer]{hinder2024onePartB}
Fabian Hinder, Valerie Vaquet, and Barbara Hammer.
\newblock One or two things we know about concept drift---a survey on
  monitoring in evolving environments. {Part B}: locating and explaining
  concept drift.
\newblock \emph{Frontiers in Artificial Intelligence}, 7:\penalty0 1330258,
  2024{\natexlab{b}}.
\newblock \doi{10.3389/frai.2024.1330258}.

\bibitem[Jensen et~al.(2007)Jensen, Ellis, Christensen, and Jensen]{Jensen2007}
J.~H. Jensen, D.~P.~W. Ellis, M.~G. Christensen, and S.~H. Jensen.
\newblock Evaluation of distance measures between {G}aussian mixture models of
  {MFCC}s.
\newblock In \emph{International Conference on Music Information Retrieval
  (ISMIR)}, pages 107--108, 2007.

\bibitem[Kampa et~al.(2011)Kampa, Hasanbelliu, and Principe]{Kampa2011}
Kittipat Kampa, Erion Hasanbelliu, and Jose~C. Principe.
\newblock Closed-form {C}auchy-{S}chwarz {PDF} divergence for mixture of
  {G}aussians.
\newblock In \emph{International Joint Conference on Neural Networks (IJCNN)},
  pages 2578--2585, 2011.

\bibitem[K{\"u}bler et~al.(2022)K{\"u}bler, Stimper, Buchholz, Muandet, and
  Sch{\"o}lkopf]{kubler2022automl}
Jonas~M. K{\"u}bler, Vincent Stimper, Simon Buchholz, Krikamol Muandet, and
  Bernhard Sch{\"o}lkopf.
\newblock {AutoML} two-sample test.
\newblock In \emph{Advances in Neural Information Processing Systems 35
  (NeurIPS)}, 2022.

\bibitem[Kuncheva(2013)]{kuncheva2013spll}
Ludmila~I. Kuncheva.
\newblock Change detection in streaming multivariate data using likelihood
  detectors.
\newblock \emph{IEEE Transactions on Knowledge and Data Engineering},
  25\penalty0 (5):\penalty0 1175--1180, 2013.
\newblock \doi{10.1109/TKDE.2011.226}.

\bibitem[Li et~al.(2020)Li, Zhao, Botta, Ionescu, and Hu]{Li2020COPOD}
Zheng Li, Yue Zhao, Nicola Botta, Cezar Ionescu, and Xiyang Hu.
\newblock {COPOD}: Copula-based outlier detection.
\newblock In \emph{IEEE International Conference on Data Mining (ICDM)}, pages
  1118--1123, 2020.

\bibitem[Li et~al.(2022)Li, Zhao, Hu, Botta, Ionescu, and Chen]{Li2022ECOD}
Zheng Li, Yue Zhao, Xiyang Hu, Nicola Botta, Cezar Ionescu, and George~H. Chen.
\newblock {ECOD}: Unsupervised outlier detection using empirical cumulative
  distribution functions.
\newblock \emph{IEEE Transactions on Knowledge and Data Engineering}, 2022.

\bibitem[Liu et~al.(2021)Liu, Lu, and Zhang]{liu2021eikmeans}
Anjin Liu, Jie Lu, and Guangquan Zhang.
\newblock Concept drift detection via equal intensity $k$-means space
  partitioning.
\newblock \emph{IEEE Transactions on Cybernetics}, 51\penalty0 (6):\penalty0
  3198--3211, 2021.
\newblock \doi{10.1109/TCYB.2020.2983962}.

\bibitem[Liu et~al.(2023)Liu, Lu, Song, Xuan, and Zhang]{liu2023cdddi}
Anjin Liu, Jie Lu, Yiliao Song, Junyu Xuan, and Guangquan Zhang.
\newblock Concept drift detection delay index.
\newblock \emph{IEEE Transactions on Knowledge and Data Engineering},
  35:\penalty0 4585--4597, 2023.
\newblock \doi{10.1109/TKDE.2022.3153349}.

\bibitem[Liu et~al.(2008)Liu, Ting, and Zhou]{Liu2008}
Fei~Tony Liu, Kai~Ming Ting, and Zhi-Hua Zhou.
\newblock Isolation forest.
\newblock In \emph{IEEE International Conference on Data Mining (ICDM)}, pages
  413--422, 2008.

\bibitem[Liu et~al.(2020)Liu, Xu, Lu, Zhang, Gretton, and
  Sutherland]{liu2020deepkernel}
Feng Liu, Wenkai Xu, Jie Lu, Guangquan Zhang, Arthur Gretton, and Danica~J.
  Sutherland.
\newblock Learning deep kernels for non-parametric two-sample tests.
\newblock In \emph{Proceedings of the 37th International Conference on Machine
  Learning (ICML)}, volume 119 of \emph{Proceedings of Machine Learning
  Research}, pages 6316--6326, 2020.

\bibitem[Lopez-Paz and Oquab(2017)]{lopezpaz2017revisiting}
David Lopez-Paz and Maxime Oquab.
\newblock Revisiting classifier two-sample tests.
\newblock In \emph{5th International Conference on Learning Representations
  (ICLR)}, 2017.

\bibitem[Lu et~al.(2019)Lu, Liu, Dong, Gu, Gama, and Zhang]{lu2019learning}
Jie Lu, Anjin Liu, Fan Dong, Feng Gu, Jo{\~a}o Gama, and Guangquan Zhang.
\newblock Learning under concept drift: A review.
\newblock \emph{IEEE Transactions on Knowledge and Data Engineering},
  31\penalty0 (12):\penalty0 2346--2363, 2019.
\newblock \doi{10.1109/TKDE.2018.2876857}.

\bibitem[Lukats et~al.(2025)Lukats, Zielinski, Hahn, and
  Stahl]{lukats2025benchmark}
Daniel Lukats, Oliver Zielinski, Axel Hahn, and Frederic~T. Stahl.
\newblock A benchmark and survey of fully unsupervised concept drift detectors
  on real-world data streams.
\newblock \emph{International Journal of Data Science and Analytics},
  19\penalty0 (1):\penalty0 1--31, 2025.
\newblock \doi{10.1007/s41060-024-00620-y}.

\bibitem[Pedregosa et~al.(2011)]{Pedregosa2011}
F.~Pedregosa et~al.
\newblock Scikit-learn: Machine learning in {P}ython.
\newblock \emph{Journal of Machine Learning Research}, 12:\penalty0 2825--2830,
  2011.

\bibitem[Pelosi et~al.(2025)Pelosi, Cacciagrano, and Piangerelli]{Pelosi2025}
Daniele Pelosi, Diletta Cacciagrano, and Marco Piangerelli.
\newblock Explainability and interpretability in concept and data drift: A
  systematic literature review.
\newblock \emph{Algorithms}, 18\penalty0 (7):\penalty0 443, 2025.
\newblock \doi{10.3390/a18070443}.

\bibitem[Pimentel et~al.(2014)Pimentel, Clifton, Clifton, and
  Tarassenko]{Pimentel2014}
Marco~A.F. Pimentel, David~A. Clifton, Lei Clifton, and Lionel Tarassenko.
\newblock A review of novelty detection.
\newblock \emph{Signal Processing}, 99:\penalty0 215--249, 2014.

\bibitem[Rabanser et~al.(2019)Rabanser, G{\"u}nnemann, and
  Lipton]{rabanser2019failing}
Stephan Rabanser, Stephan G{\"u}nnemann, and Zachary~C. Lipton.
\newblock Failing loudly: An empirical study of methods for detecting dataset
  shift.
\newblock In \emph{Advances in Neural Information Processing Systems 32
  (NeurIPS)}, pages 1394--1406, 2019.

\bibitem[Rabin et~al.(2011)Rabin, Peyr{\'e}, Delon, and
  Bernot]{rabin2011sliced}
Julien Rabin, Gabriel Peyr{\'e}, Julie Delon, and Marc Bernot.
\newblock Wasserstein barycenter and its application to texture mixing.
\newblock In \emph{Scale Space and Variational Methods in Computer Vision
  (SSVM)}, volume 6667 of \emph{Lecture Notes in Computer Science}, pages
  435--446. Springer, 2011.

\bibitem[Rayana(2016)]{ODDS2016}
Shebuti Rayana.
\newblock {ODDS} library, 2016.
\newblock \url{http://odds.cs.stonybrook.edu}.

\bibitem[Reynolds(2009)]{Reynolds2009}
Douglas~A. Reynolds.
\newblock Gaussian mixture models.
\newblock \emph{Encyclopedia of Biometrics}, pages 659--663, 2009.

\bibitem[Sakurada and Yairi(2014)]{Sakurada2014}
Mayu Sakurada and Takehisa Yairi.
\newblock Anomaly detection using autoencoders with nonlinear dimensionality
  reduction.
\newblock \emph{Proc. MLSDA Workshop on Machine Learning for Sensory Data
  Analysis}, pages 4--11, 2014.

\bibitem[Sch{\"o}lkopf et~al.(2001)Sch{\"o}lkopf, Platt, Shawe-Taylor, Smola,
  and Williamson]{Scholkopf2001}
Bernhard Sch{\"o}lkopf, John~C. Platt, John Shawe-Taylor, Alex~J. Smola, and
  Robert~C. Williamson.
\newblock Estimating the support of a high-dimensional distribution.
\newblock \emph{Neural Computation}, 13\penalty0 (7):\penalty0 1443--1471,
  2001.

\bibitem[Schrab et~al.(2023)Schrab, Kim, Albert, Laurent, Guedj, and
  Gretton]{schrab2023mmdagg}
Antonin Schrab, Ilmun Kim, M{\'e}lisande Albert, B{\'e}atrice Laurent, Benjamin
  Guedj, and Arthur Gretton.
\newblock {MMD} aggregated two-sample test.
\newblock \emph{Journal of Machine Learning Research}, 24\penalty0
  (194):\penalty0 1--81, 2023.

\bibitem[Schwarz(1978)]{Schwarz1978}
Gideon Schwarz.
\newblock Estimating the dimension of a model.
\newblock \emph{The Annals of Statistics}, 6\penalty0 (2):\penalty0 461--464,
  1978.

\bibitem[Shewhart(1931)]{shewhart1931economic}
Walter~A. Shewhart.
\newblock \emph{Economic Control of Quality of Manufactured Product}.
\newblock D. Van Nostrand Company, New York, 1931.

\bibitem[Siffer et~al.(2017)Siffer, Fouque, Termier, and
  Largou{\"e}t]{Siffer2017}
Alban Siffer, Pierre-Alain Fouque, Alexandre Termier, and Christine
  Largou{\"e}t.
\newblock Anomaly detection in streams with extreme value theory.
\newblock In \emph{ACM SIGKDD International Conference on Knowledge Discovery
  and Data Mining}, pages 1067--1075, 2017.

\bibitem[Sugiyama et~al.(2013)Sugiyama, Kanamori, Suzuki, du~Plessis, Liu, and
  Takeuchi]{sugiyama2013lsdd}
Masashi Sugiyama, Takafumi Kanamori, Taiji Suzuki, Marthinus~Christoffel
  du~Plessis, Song Liu, and Ichiro Takeuchi.
\newblock Density-difference estimation.
\newblock \emph{Neural Computation}, 25\penalty0 (10):\penalty0 2734--2775,
  2013.
\newblock \doi{10.1162/NECO_a_00492}.

\bibitem[Tavallaee et~al.(2009)Tavallaee, Bagheri, Lu, and
  Ghorbani]{Tavallaee2009}
Mahbod Tavallaee, Ebrahim Bagheri, Wei Lu, and Ali~A. Ghorbani.
\newblock A detailed analysis of the {KDD} {CUP} 99 data set.
\newblock In \emph{IEEE Symposium on Computational Intelligence for Security
  and Defense Applications}, 2009.

\bibitem[Wan et~al.(2024)Wan, Liang, and Yoon]{wan2024mcddd}
Ke~Wan, Yi~Liang, and Susik Yoon.
\newblock Online drift detection with maximum concept discrepancy.
\newblock In \emph{Proceedings of the 30th ACM SIGKDD Conference on Knowledge
  Discovery and Data Mining (KDD)}, pages 2924--2935, 2024.
\newblock \doi{10.1145/3637528.3672016}.

\bibitem[Zhao et~al.(2019)Zhao, Nasrullah, and Li]{Zhao2019PyOD}
Yue Zhao, Zain Nasrullah, and Zheng Li.
\newblock {PyOD}: A {P}ython toolbox for scalable outlier detection.
\newblock \emph{Journal of Machine Learning Research}, 20\penalty0
  (96):\penalty0 1--7, 2019.

\end{thebibliography}

\end{document}